\documentclass[journal]{IEEEtran}

\usepackage{epsfig}
\usepackage{graphicx}
\usepackage{subfigure}
\usepackage{cite}
\usepackage{bm}
\usepackage{float}
\usepackage{algorithm}
\usepackage{algpseudocode}
\usepackage{verbatim}
\usepackage{dsfont}
\usepackage{array}
\usepackage{booktabs}
\usepackage{tabu}
\usepackage{makecell}
\usepackage{setspace}
\usepackage{amsmath}
\usepackage{multirow}
\usepackage{url}
\usepackage{amsfonts}
\usepackage{multirow}
\usepackage{hyperref}

\usepackage[figuresright]{rotating}
\usepackage[section]{placeins}
\UseRawInputEncoding

\begin{document}

\title{Transformers Meet Visual Learning Understanding: A Comprehensive Review}

\author{
Yuting Yang,~\IEEEmembership{Student~Member,~IEEE,}
Licheng Jiao,~\IEEEmembership{Fellow,~IEEE,}\\
Xu Liu,~\IEEEmembership{Member,~IEEE,}
Fang Liu,~\IEEEmembership{Senior Member,~IEEE,}\\
Shuyuan Yang,~\IEEEmembership{Senior Member,~IEEE,}
Zhixi Feng,~\IEEEmembership{Member,~IEEE,}
Xu Tang,~\IEEEmembership{Member,~IEEE}


\thanks{This work was supported in part by
the Key Scientific Technological Innovation Research Project by Ministry of Education,
the State Key Program and the Foundation for Innovative Research Groups of the National Natural Science Foundation of China (61836009, 61621005, 62076192),
Key Research and Development Program in Shaanxi Province of China (2019ZDLGY03-06),
the Major Research Plan of the National Natural Science Foundation of China (91438201, 91438103, and 61801124),
the National Natural Science Foundation of China (U1701267, 62006177, 61871310, 61902298, 61573267, 91838303 and 61906150),
the Fund for Foreign Scholars in University Research and Teaching Program’s 111 Project (B07048),
the Program for Cheung Kong Scholars and Innovative Research Team in University (IRT 15R53),
the ST Innovation Project from the Chinese Ministry of Education,
the National Science Basic Research Plan in Shaanxi Province of China(2019JQ-659),
the Scientific Research Project of Education Department In Shaanxi Province of China (No.20JY023),
the fundamental research funds for the central universities (XJS201901, XJS201903, JBF201905, JB211908),
and the CAAI-Huawei MindSpore Open Fund.\textit{(Corresponding author: Licheng Jiao.)}}

\thanks{The authors are with the Key Laboratory of Intelligent Perception and Image Understanding of the Ministry of Education of China, International Research Center of Intelligent Perception and Computation, School of Artificial Intelligence, Xidian University, Xi'an, China (e-mail: lchjiao@mail.xidian.edu.cn; ytyang\_1@stu.xidian.edu.cn).}}

\markboth{Transformers Meet Visual Learning Understanding: A Comprehensive Review}%
{Shell \MakeLowercase{\textit{et al.}}: Bare Demo of IEEEtran.cls for Journals}

\maketitle

\begin{abstract}
Dynamic attention mechanism and global modeling ability make Transformer show strong feature learning ability. In recent years, Transformer has become comparable to CNNs methods in computer vision. This review mainly investigates the current research progress of Transformer in image and video applications, which makes a comprehensive overview of Transformer in visual learning understanding. First, the attention mechanism is reviewed, which plays an essential part in Transformer. And then, the visual Transformer model and the principle of each module are introduced. Thirdly, the existing Transformer-based models are investigated, and their performance is compared in visual learning understanding applications. Three image tasks and two video tasks of computer vision are investigated. The former mainly includes image classification, object detection, and image segmentation. The latter contains object tracking and video classification. It is significant for comparing different models' performance in various tasks on several public benchmark data sets. Finally, ten general problems are summarized, and the developing prospects of the visual Transformer are given in this review.
\end{abstract}

\IEEEpeerreviewmaketitle
\section{Introduction}\label{sec:1}
\IEEEPARstart{D}{eep} learning \cite{lecun2015deep} has been developed rapidly, and convolutional neural networks (CNNs) have occupied a dominant position in various fields of deep learning \cite{mahmud2018applications,zhao2019object}. However, Transformer \cite{vaswani2017attention} has gradually broken this situation in recent years. It abandoned CNNs and RNNs used in previous deep learning tasks and made breakthroughs in natural language processing (NLP), computer vision (CV), and other fields. Gradually, Transformer-based models have been well developed in recent three years.

The initial Transformer model is formally proposed in the paper named ``Attention is all you need'' in 2017 \cite{vaswani2017attention}. It comes from the machine translation model seq2seq \cite{sutskever2014sequence} in NLP. Furthermore, the encoder-decoder architecture is also adopted in the Transformer model. It mainly evolves from an attention module, self-attention, one of the existing attention models. As for attention mechanisms, many attention models have appeared to improve the recognition result. The existing attention models mainly include channel attention, spatial attention, and self-attention \cite{guo2021attention}. The core of Transformer is self-attention.

In the beginning, Transformer is a novel method that shows great success in NLP. Later, it has been extended to different tasks in CV, such as high-resolution image synthesis \cite{dalmaz2021resvit}, object tracking \cite{DBLP:journals/corr/abs-2101-02702}, object detection \cite{jiao2019survey,carion2020end,qian2019oriented}, classification \cite{dey2008rough}, segmentation \cite{zheng2021rethinking}, image processing \cite{9580642}, re-identification \cite{li2021exploiting,luo2020stnreid,9674853} and so on.

In the past three years, Transformer has evolved a series of variants, also known as X-Transformer \cite{lin2021survey}. Various Transformers came into being and made good application progress in various tasks. The research shows that the pre-trained Transformer model achieves the state-of-art in various tasks. The effect of the Transformer model is remarkable, especially in the ImageNet classification task. ViT \cite{dosovitskiy2020image}, BoTNet \cite{srinivas2021bottleneck} and Swin Transformer \cite{liu2021swin} have been proposed one after another, and have achieved performance breakthroughs time and time again. This review attracts our attention to the developing progress on Transformer in image and video applications of visual learning understanding.
\begin{table*}[]
\caption{The related surveys and their main contents referenced in the surveys. The citation statistics is counted up to January 10, 2022}
\footnotesize
\label{tab:surveys}
\begin{tabular}{c|c|ccccccc|c}
\toprule
\multirow{3}{*}{\textbf{Titles}}           & \multirow{3}{*}{\textbf{Citation}} & \multicolumn{7}{c|}{\textbf{Contents}}                                                                                                                                                                                                                               & \multirow{3}{*}{\textbf{\begin{tabular}[c]{@{}c@{}}Issues and \\ prospects\end{tabular}}} \\ \cline{3-9}
                                   &                           & \multicolumn{1}{c|}{\multirow{2}{*}{attention}} & \multicolumn{1}{c|}{\multirow{2}{*}{module}} & \multicolumn{3}{c|}{Image}                                                                               & \multicolumn{2}{c|}{Video}                     &                                                                                  \\ \cline{5-9}
                                   &                           & \multicolumn{1}{c|}{}                            & \multicolumn{1}{c|}{}                        & \multicolumn{1}{c|}{classification} & \multicolumn{1}{c|}{segmentation} & \multicolumn{1}{c|}{detection} & \multicolumn{1}{c|}{tracking} & classifcaition &                                                                                  \\ \toprule
Khan S, et al. \cite{lin2021survey}& 172                       & \multicolumn{1}{c|}{Y}                           & \multicolumn{1}{c|}{Y}                       & \multicolumn{1}{c|}{Y}              & \multicolumn{1}{c|}{Y}            & \multicolumn{1}{c|}{Y}         & \multicolumn{1}{c|}{Y}        & N              & N                                                                                \\ \hline
Han K, et al. \cite{han2020survey}       & 108                       & \multicolumn{1}{c|}{Y}                           & \multicolumn{1}{c|}{Y}                       & \multicolumn{1}{c|}{Y}              & \multicolumn{1}{c|}{Y}            & \multicolumn{1}{c|}{Y}         & \multicolumn{1}{c|}{N}        & N              & N                                                                                \\ \hline
Correia A S, \cite{correia2021attention}       & 4                         & \multicolumn{1}{c|}{Y}                           & \multicolumn{1}{c|}{N}                       & \multicolumn{1}{c|}{Y}              & \multicolumn{1}{c|}{Y}            & \multicolumn{1}{c|}{Y}         & \multicolumn{1}{c|}{Y}        & N              & Y                                                                                \\ \hline
Tay Y, et al. \cite{tay2020efficient} & 191                       & \multicolumn{1}{c|}{N}                           & \multicolumn{1}{c|}{Y}                       & \multicolumn{1}{c|}{N}              & \multicolumn{1}{c|}{N}            & \multicolumn{1}{c|}{N}         & \multicolumn{1}{c|}{N}        & N              & Y                                                                                \\ \hline
Khan S, et al. \cite{khan2021transformers}  & 172                       & \multicolumn{1}{c|}{Y}                           & \multicolumn{1}{c|}{Y}                       & \multicolumn{1}{c|}{Y}              & \multicolumn{1}{c|}{Y}            & \multicolumn{1}{c|}{Y}         & \multicolumn{1}{c|}{N}        & N              & N                                                                                \\ \hline
Brasoveanu, et al. \cite{bracsoveanu2020visualizing}       & 8                         & \multicolumn{1}{c|}{N}                           & \multicolumn{1}{c|}{N}                       & \multicolumn{1}{c|}{N}              & \multicolumn{1}{c|}{N}            & \multicolumn{1}{c|}{N}         & \multicolumn{1}{c|}{N}        & N              & Y                                                                                \\ \hline
Ours                               & /                         & \multicolumn{1}{c|}{Y}                           & \multicolumn{1}{c|}{Y}                       & \multicolumn{1}{c|}{Y}              & \multicolumn{1}{c|}{Y}            & \multicolumn{1}{c|}{Y}         & \multicolumn{1}{c|}{Y}        & Y              & Y                                                                                \\ \hline
\end{tabular}
\end{table*}

Compared with the related surveys \cite{lin2021survey,han2020survey,correia2021attention,tay2020efficient,bracsoveanu2020visualizing}, this review conducts a comprehensive investigation from the model mechanism of Transformer, the application progress of visual learning understanding applications, and the performance comparison of various models on the public benchmarks. The related surveys and their main contents are investigated in TABLE \ref{tab:surveys}. This review aims to give readers a comprehensive understanding of the Transformer, its principle, and existing application progress. In addition, it provides an experimental comparison for the investigated image and video researches. Meanwhile, it provides further ideas for deep learning researchers.

The main contributions of this survey are shown as follows:

1) Transformer-based methods for visual learning understanding are comprehensively investigated, and some remarks are given.

2) The attention mechanism has been reviewed, which plays an essential part in Transformer.

3) Each part of the original visual Transformer model is detailed. It is essential to understand the principle of visual Transformers fully.

4) The application progress of Transformer-based models is summarized in visual learning understanding, including image classification, target tracking, image segmentation, target tracking, and video classification. And then, the performance comparison of each model is given in each subsection, which provides an experimental comparison for related researchers.

5) Ten public issues of Transformer are summarized. It will provide researchers with further research ideas.

The rest of this paper is organized from the following aspects. Section \ref{sec:2} has a review of the attention mechanism.
Section \ref{sec:3} details the initial visual Transformer modules.
Section \ref{sec:4} presents the research progress of Transformer in the image tasks of visual learning understanding.
Section \ref{sec:5} presents the research progress of Transformer in video tasks of visual learning understanding.
Section \ref{sec:6} mainly summarizes ten public challenges of Transformer and gives the conclusion of this review.

\section{Attention mechanism}\label{sec:2}
Attention mechanism was proposed in the 1990s \cite{chaudhari2021attentive}.
It refers to applying human perception and attention behavior to the machine, which can learn to perceive the important and unimportant parts of the data. In CV, the attention mechanism lets the machine perceive the target information in the image and suppress the image's background information.
Introducing attention mechanisms can alleviate the limitation of computational power and optimization algorithms in deep learning.

\begin{figure}
  \centering
  \includegraphics[width=5.0cm]{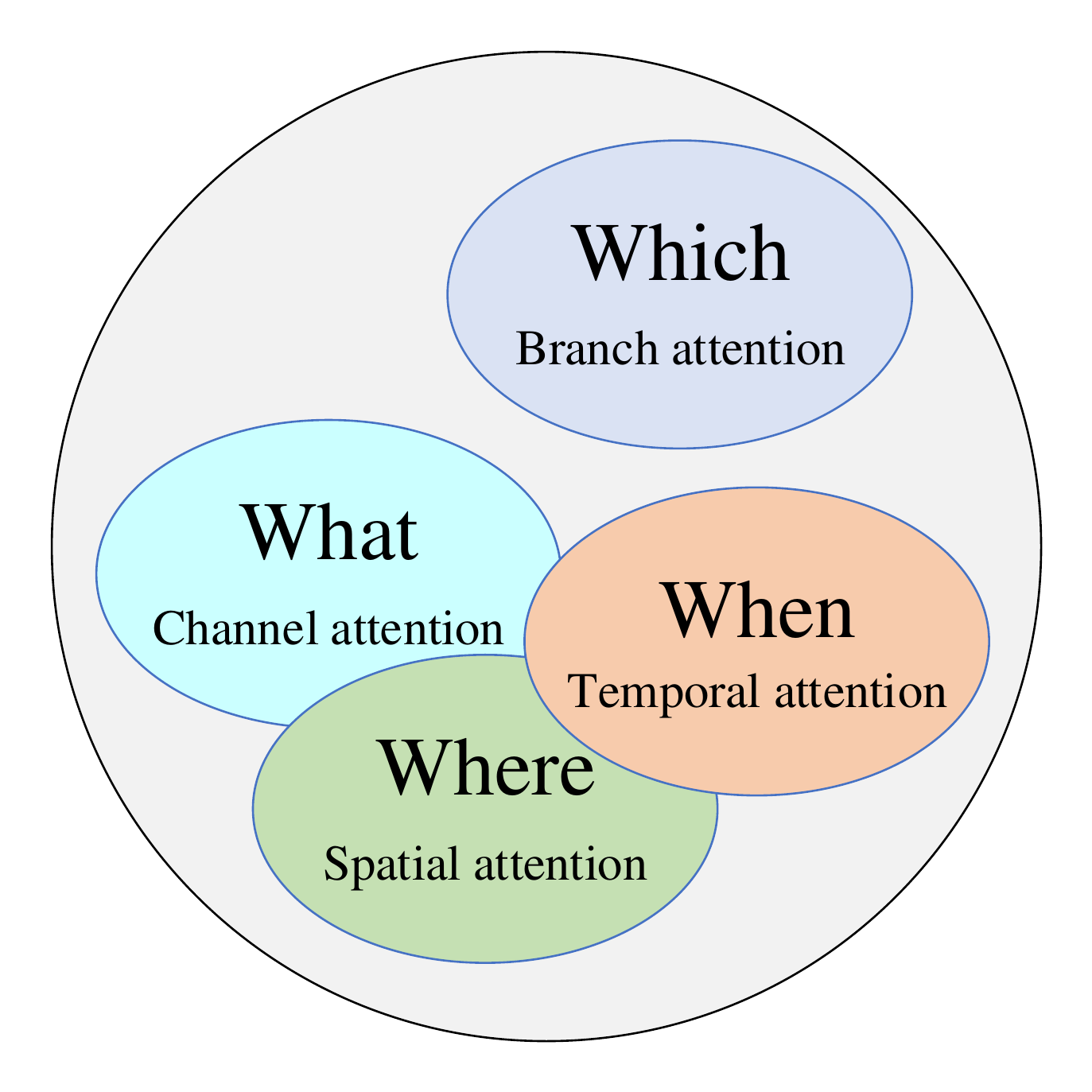}\\
  \caption{Attention mechanisms can be categorised according to data domain. It mainly includes four types of attention mechanisms: channel attention, spatial attention, temporal attention, and branch attention. Their attention are paid on what, where, when, and which to attention respectively.}
  \label{fig:1}
\end{figure}

The existing attention mechanisms in deep learning are classified according to different angles \cite{hashemi2015survey}.
Whether all hidden states of the encoder are considered during decoding, it is divided into global and local attention mechanisms.
From the perspective of the attention domain, it can be divided into attention domain, spatial domain, channel domain, and the mixed domain.
According to different coding methods, it can be divided into soft attention mechanism, hard attention mechanism, and self-attention mechanism.
Among them, self-attention is the research core of the Transformer model.
There are mainly four attention mechanisms commonly used in CV (shown as Fig. \ref{fig:1}),
including channel attention \cite{hu2018squeeze}, spatial attention \cite{guo2021attention}, temporal attention \cite{liu2021tam} and branch attention \cite{srivastava2015highway}.

\subsection{Channel attention}
Channel attention is first proposed by Hu et al. in \cite{hu2018squeeze}. It is a mechanism determining which channel to pay attention to. It displays the correlation (feature map) between different channels through modeling and automatically obtains the importance of each feature channel through network learning. Finally, different weight coefficients are assigned to each channel to enhance essential features and suppress non-important features.
SENet is a typical channel attention model. It mainly contains the squeeze and excitation steps. Later on, some improved models appearing, including GSoP-Net \cite{gao2019global}, FcaNet \cite{qin2021fcanet}, SRM \cite{lee2019srm}, GCT \cite{yang2020gated}, and so on.
%


\subsection{Spatial attention}
Spatial attention is proposed for where to pay attention. Spatial attention aims to enhance the expression of critical regional features. It converts the spatial information in the original picture into another space and retains the critical information through the spatial conversion module. It generates a weighted mask for each location and weights the output to enhance the target region of interest and weaken the irrelevant background region.
There are four types of spatial attention models, including RNN-based attention (i.e, RAM \cite{mnih2014recurrent}, Hard and soft attention \cite{xu2015show}), predict the relevant region explicitly (i.e, STN \cite{jaderberg2015spatial}, DCN \cite{dai2017deformable}), predict the relevant region implicitly (i.e, GENet \cite{hu2018gather}, PSANet \cite{zhao2018psanet}), and models based on self-attention \cite{guo2021attention} (i.e, SASA \cite{ramachandran2019stand}, ViT \cite{dosovitskiy2020image}).

%


\subsection{Temporal attention}
Temporal attention is a mechanism determining when to pay attention. It is often used in video processing or analysis. There are several types of models, including combining local and global attention and self-attention based. The typical models contain GLTR \cite{li2019global} and TAM \cite{liu2021tam}. The self-attention-based GLTR model is used for ReID tasks, and TAM combines local and global attention for behavior recognition tasks.

\subsection{Branch attention}
Branch attention is a dynamic branch selection mechanism. It decides which branch to pay attention to through a multi-branch structure. Typical branch attention includes the highway network \cite{srivastava2015highway} and SKNet \cite{li2019selective} and CondConv \cite{yang2019condconv}. The highway network and SKNet mainly combine different branches, and the CondConv mainly combine different convolution kernels.

\subsection{Remarks}
Attention mechanism has been a more critical research direction in CV. The core of Transformer is self-attention, which originated in NLP. Comparing different attention mechanisms gives readers a preliminary understanding of different attention mechanisms. The self-attention module plays an essential role in Transformer. Combining other attention models into Transformer models or introducing the visual attention thoughts into self-attention may be a novel research direction.

\section{Transformer modules}\label{sec:3}
The structure of the original visual Transformer model mainly comprises encoding and decoding parts. There are three essential modules: attention, position-wise feed-forward network, and position encoding.
\subsection{Attention in Transformer}
\begin{figure}[htp]
\begin{center}
   \includegraphics[width=0.9\linewidth]{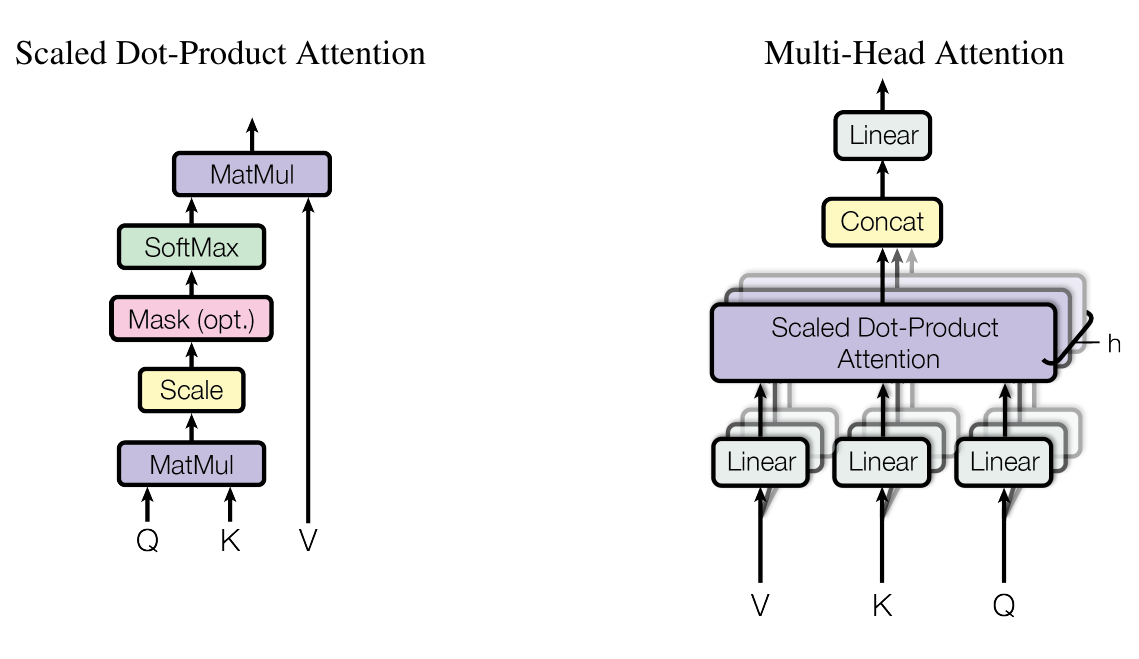}
\end{center}
   \caption{Scaled attention and the multi-head attention.}
\label{fig:scaledattentionl}
\end{figure}

The Transformer takes the attention mechanism as the core, which mainly involves three different attention modules. It includes self-attention, mask attention and cross-sequence attention \cite{lin2021survey}. Fig. \ref{fig:scaledattentionl} shows the scale dot-product attention and multi-head attention involved in Transformer.

\subsubsection{Self-attention}
In 2017, Ashish Vaswani et al. put forward visual Transformer \cite{vaswani2017attention}.
Subsequently, the self-attention mechanism is widely designed for visual learning understanding.
It aims to capture the internal relevance of data or features, reducing reliance on external information.
It can solves the long-distance dependence problem by calculating the mutual influence between different patches of the image.

For an image $X$, the self-attention can be modeled as follows. The queries(Q), keys(K), and values(V) can be obtained by the transformation of the input. A common form of $Q$, $K$, and $V$ can be formulated as Eq. (\ref{eq:qkv}).
\begin{equation}
K=W^{K}X,
Q=W^{Q}X,
V=W^{V}X.
\label{eq:qkv}
\end{equation}

And then, the scaled dot-product attention can be expressed as Eq. (\ref{eq:att}).
\begin{equation}
Attention(Q,K,V)=softmax(\frac{QK^{T}}{\sqrt{d_{k}}})V,
\label{eq:att}
\end{equation}
where the $(\frac{QK^{T}}{\sqrt{d_{k}}})$ is named as the attention matrix.
\begin{figure*}[]
\setlength{\belowcaptionskip}{-0.1cm}
\begin{center}
   \includegraphics[width=0.9\linewidth]{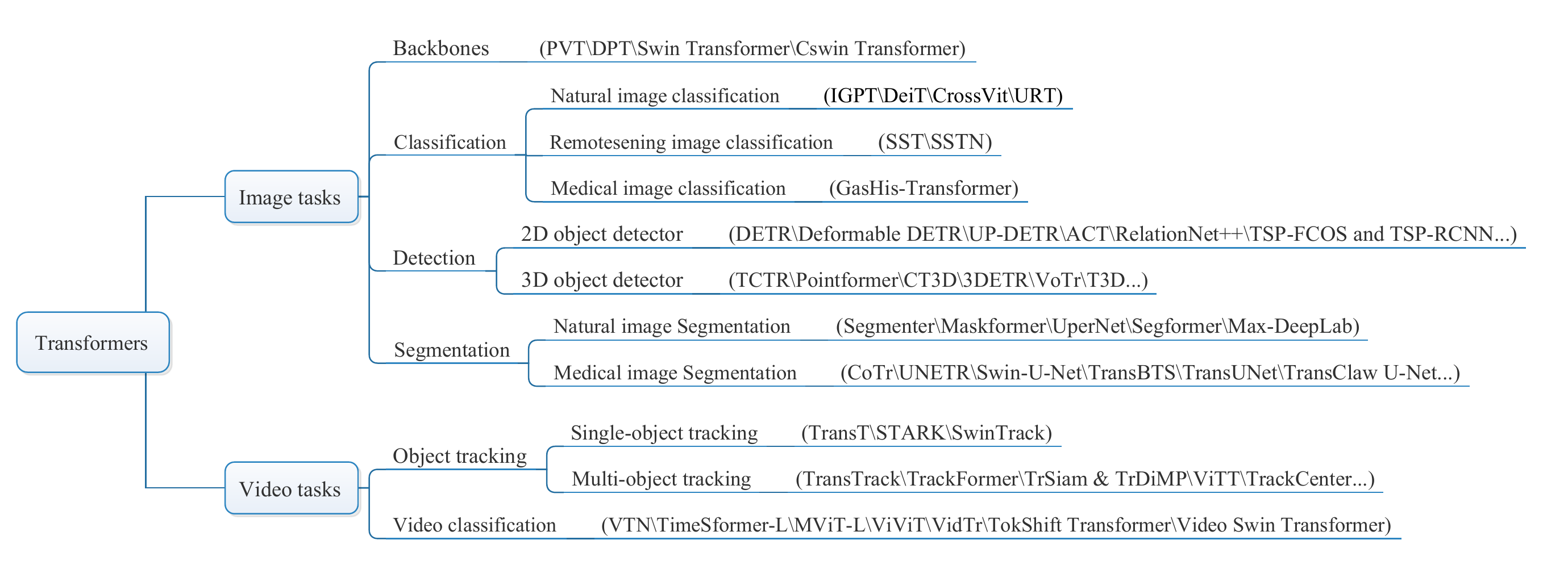}
\end{center}
   \caption{The framework of Transformers for visual learning and understanding. The backbone, image classification, object detection, image segmentation based on Transformer are mainly investigated for image tasks. The object tracking, video classification based on Transformer for video tasks are reviewed for video tasks.}
\label{fig:paperframe}
\end{figure*}
\subsubsection{Multi-head attention}
The right model shown in Fig. \ref{fig:scaledattentionl} is called multi-head attention (MHA). It can be formulated as Eq. (\ref{eq:mha}).
\begin{equation}
MultiHead(Q,K,V)=Concat(head_{1},\ldots,head_{h})W^{O}),
\label{eq:mha}
\end{equation}
where $head_{i}=Attention(W_{i}^{Q}X,W_{i}^{K}X,W_{i}^{V}X)$.

\subsection{Feedforward network}
In addition to the attention mechanism, the original Transformer model also includes a fully connected feedforward network (FFN) and position embedding (PE). FFN is mainly composed of two linear transformations, with a ReLU activation function in between. Its output can be expressed as the following Eq. (\ref{eq:ffn}).
\begin{equation}
FFN(x)=max(0,xW_{1}+b_{1})W_{2}+b_{2}.
\label{eq:ffn}
\end{equation}
\subsection{Position encoding}
In Transformer, sine and cosine functions are mainly used for position encoding. The specific coding method is formulated as Eq. \ref{eq:pos}.
\begin{equation}
\begin{aligned}
PE(pos,2i)=sin(pos/10000^{2i/d_{model}}),\\
PE(pos,2i+1)=cos(pos/10000^{2i/d_{model}}),
\label{eq:pos}
\end{aligned}
\end{equation}
where $pos$ represents the position, and $i$ means the dimension. Each dimension of the position code corresponds to a sine curve.
\subsection{Complexity analysis}
\textbf{The complexity analysis of self-attention.}
The computational complexity for an image $A$, the corresponding $Q$, $K$, and $V$ are all of $n \times d$ dimension.
The similarity calculation is formulated as $QK^{T}$, where the matrix operation of $n\times d$ and $d\times n$,
the $n \times n$ matrix is obtained. Thus its computational complexity is $O(n^{2}d)$.
The following softmax calculation is performed on each row, and its complexity is $O(n)$. Then the complexity of $n$ rows is $O(n^{2})$. The weighted summation part is the operation of the $n\times n$ matrix and the $n\times d$ matrix to obtain the $n\times d$ matrix with a complexity of $O(n^{2}d)$. Therefore, the time complexity of the self-attention module is $O(n^{2}d)$.

\textbf{The complexity analysis of multi-head attention.}
As the multi-head attention (MHA) introduced in the Eq. (\ref{eq:mha}), assuming there are $h$ heads, each head needs to map three matrices to the $d^{q}, d^{k}, d^{v}$ dimension.
The linear mapping complexity of input is equal to $n\times d$ and $d\times d/h$ operations (ignoring constant coefficients), thus its complexity is $O(nd^{2})$.
Attention operation complexity mainly depends on the similarity calculation and weighted average cost, $n\times d/h$ and $d/h\times n$ operations, the computational complexity is $O(n^{2}d)$.
Output linear mapping complexity: concatenation operations are spliced together to form an $n\times d$ matrix,
and then undergo linear output linear mapping to ensure that the input and output are the same.
Therefore, the complexity of calculating $n\times d$ and $d\times d$ is $O(nd^{2})$. Thus, the final complexity of MHA is $O(n^{2}d+nd^{2})$.
\subsection{Remarks}
ViT gives the most primitive visual Transformer, which contains the main modules above.
Many existing Transformer-based methods are also improved from these aspects,
including improving block, improving position coding, improving Encoder, adding Decoder, etc.
For example, \cite{yue2021vision} mainly improves the block tokens. It enhances the image block in ViT. The model updates the sampling position iteratively. In each iteration, the image block of the current sampling step is fed to the Transformer encoding layer, and a set of sampling offsets are predicted to update the sampling location for the next step.

The difference between absolute and relative position-coding in CV is not obvious.
Four relative position-coding methods are proposed in \cite{wu2021rethinking}, which is easy to inserted into other networks and can reduce the computational complexity. Research on the improvement of the coding part has also emerged in large numbers, such as expression recognition task \cite{xue2021transfer}, semantic segmentation task \cite{xie2021segmenting}, target detection task \cite{wang2021pnp} and so on. Improvements on the decoding part are also in progress. For example, \cite{li2021improved} is inspired by pooling attention for the three different downstream tasks of image classification, target detection, and video classification and adjusts the fixed-resolution input to have a multiplicity from high-resolution to low-resolution (the feature hierarchy of each stage).

\section{Transformer for image tasks}\label{sec:4}
Image classification, object detection, and image segmentation are three basic image tasks in CV.
Transformer-based methods for these three tasks have been well developed.
There are Transformer-based backbones and Transformer-based necks.
The formers are evaluated on all three tasks, and the latter are often assessed on either of them.
The related Transformer models and their corresponding experimental results are investigated.
\subsection{Backbones}
Simple and powerful backbones are a major focus of Transformer-related research. They incorporate self-attention into a variety of computer vision tasks. Here, we introduce the related Transformer backbones, including Swin Transformer \cite{liu2021swin}, CSWin Transformer \cite{dong2021cswin}, PVT \cite{wang2021pyramid}, DPT \cite{ranftl2021vision}, BotNet \cite{srinivas2021bottleneck}, and CrossFormer \cite{wang2021crossformer}.

\textbf{Swin Transformer}.
Swin Transformer \cite{liu2021swin} introduces the scale changes and builds a hierarchical Transformer using the hierarchical construction method commonly used in CNN. Meanwhile, raising the local idea and the self-attention calculation for the window area without overlap effectively alleviates the computational complexity of the Transformer in image task. The proposal of Swin Transformer solves the single-scale problem of previous Transformers.

The Swin Transformer's architecture consists of 4 stages, including linear embedding, Swin Transformer block, patch partition, and patch merging.
The input image is first divided into non-overlapping patches by the patch partition. And then, the input is merged according to the $2\times 2$ adjacent patches by the patch merging through the Swin Transformer Block. Finally, it is repeatedly sent to the patch merging, Swin Transformer Block operations. The final output of the Swin Transformer is the recognition result.
Swin Transformer \cite{liu2021swin} achieves 87.3\% top-1 accuracy on ImageNet-1K, 58.7\% box AP and 51.1\% mask AP on COCO test-dev, and 53.5\% mIoU on ADE20K val.

\textbf{CSWin Transformer}.
%
CSWin Transformer \cite{dong2021cswin} mainly solves the problem of interaction domain limitation between tokens caused by the high computational cost of global attention and local self-attention. The proposed self-attention mechanism of cross-shaped windows can calculate the self-attention in the vertical and horizontal of cross-shaped windows in parallel, in which each strip is obtained by dividing the input feature into equal width strips. In addition, locally-enhanced positional encoding (LePE) is proposed to process local location information better. It can adapt to different-size input characteristics and arbitrary support input, friendly to basic downstream tasks. Combining LePE and hierarchies, CSWin Transformer with LePE shows competitive performance in downstream tasks. It achieves 85.4\% top-1 accuracy on ImageNet-1K, 53.9\% box AP and 46.4\% mask AP on the COCO, and 51.7\% mIOU on the ADE20K.

\textbf{PVT}.
Pyramid vision Transformer (PVT) \cite{wang2021pyramid} is proposed to overcome the difficulties of porting Transformer to various dense prediction tasks. It adopts a progressive shrinking pyramid to reduce the computations of large feature maps.
Besides, it inherits the advantages of both CNN and Transformer without convolutions. It can be used as a direct replacement for CNN backbones. Extensive experiments have been evaluated on different downstream tasks. PVT+RetinaNet achieves 40.4\% AP on the COCO data set, 81.7\% top-1 accuracy on the ImageNet validation set. Besides, PVT-Large+Semantic FPN shows 44.8\% mIoU on the ADE20K benchmark.

\textbf{DPT}.
DPT \cite{ranftl2021vision} is a Transformer-based visual backbone network for dense prediction tasks,
which can replace convolutional networks. It can assemble the tokens of each stage into image representation with different resolutions.
In addition, it adopts a convolutional decoder to combine different image representations into full-resolution prediction gradually.
Compared with full convolution neural networks, DPT can provide more fine-grained and global consistent predictions by representing the characteristics of the global receptive field with high-resolution processing. Experiments have been evaluated on dense prediction tasks. It shows 49.02\% mIoU on the ADE20K and 60.04\% mIoU on the Pascal Context validation set.

\textbf{BotNet}.
BotNet \cite{srinivas2021bottleneck} is presented as a backbone network for multiple computer vision tasks, which combines CNNs with self-attention. Compared with the ResNet bottleneck block, the bottleneck Transformer block only replaces the spatial $3\times 3$ convolution layer with multi-head self-attention. BotNet mainly adopts global self-attention to replace the spatial convolutions in the final three bottleneck blocks of a ResNet. It shows 84.7\% top-1 accuracy on the ImageNet. Besides, BotNet with Mask R-CNN framework achieves 44.4\% Mask AP and 49.7\% Box AP on the COCO Instance Segmentation data set.

\textbf{CrossFormer}.
Crossformer \cite{wang2021crossformer} is cross-scale attention proposed to establish the relationship between objects with large size differences in the image. It can establish the interaction between different scale features. Its core includes a cross-scale embedding layer (CEL) and long and short distance attention (LSDA). The former provides cross-scale features for the self-attention module by mixing each embedding with multiple patches of different scales. The latter mainly retains the small-scale and large-scale features of embedding. Experiments show that CrossFormer is superior to other visual converters in basic computer vision tasks. CrossFormer-L shows 84.0\% top-1 accuracy on the ImageNet. CrossFormer-RetinaNet shows 46.2\% AP on COCO2017 Val, and CrossFormer-Semantic FPN shows 51.4\% mIOU on the ADE20K validation set.

\subsection{Image classification}
\begin{figure}[htp]
\setlength{\belowcaptionskip}{-0.1cm}
\begin{center}
   \includegraphics[width=1.0\linewidth]{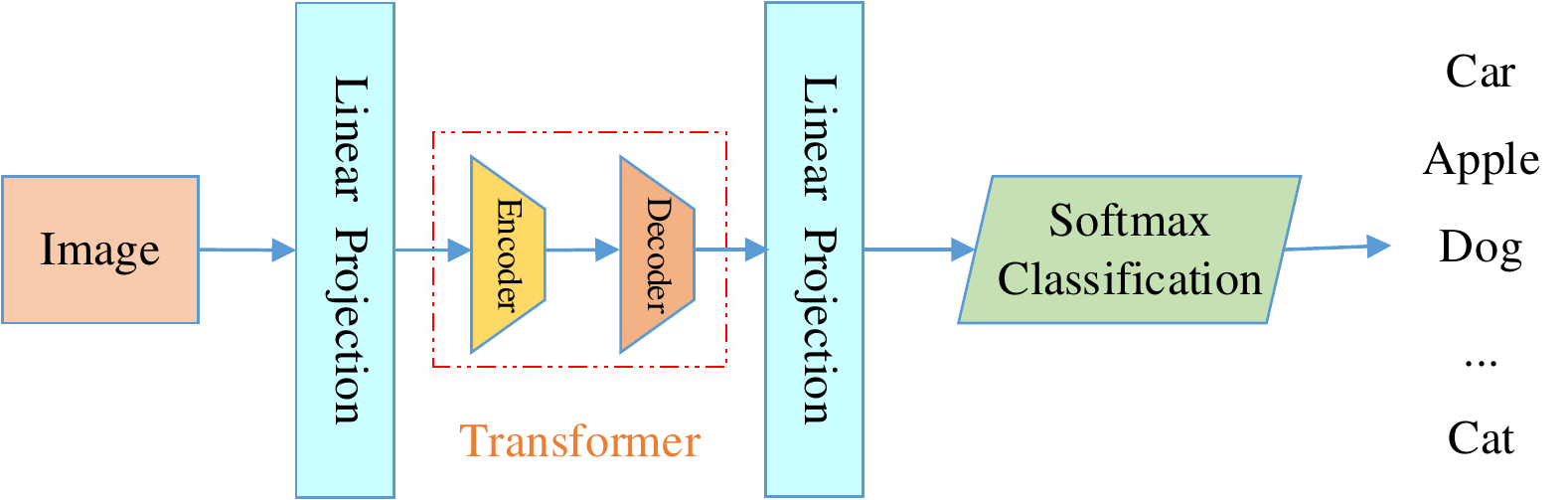}
\end{center}
   \caption{The overall framework of image classification based on Transformer.}
\label{fig:classification}
\end{figure}

The overall framework of Transformer-based methods for image classification is shown in Fig. \ref{fig:classification}.
The input image is first linearly mapped, and then is encoded and decoded by Transformer.
Subsequently, it is linearly mapped and sent to the softmax classifier to predict the image category. There are different Transformer-based models for natural, remote sensing, and medical image classification.

\subsubsection{Natural image classification}

\textbf{ViT}.
Visual Transformer \cite{dosovitskiy2020image}, namely ViT, directly applies pure Transformer to a sequence of image blocks and performs image classification tasks. It is mainly composed of position-coding embedding, linear mapping, Transformer encoder, learnable layer, and layer normalization.
First, the input image is divided into fixed-size patches. And then, a linear mapping is performed for each patch, and a position code is embedded. Subsequently, it is sent to the Transformer's encoder for encoding, and then the classification is predicted by regression through the MLP header. The ViT with pre-training can achieve better performance than the most advanced convolutional networks. And its required computing resources are relatively reduced.

\begin{figure}[htp]
\setlength{\belowcaptionskip}{-0.1cm}
\begin{center}
   \includegraphics[width=0.9\linewidth]{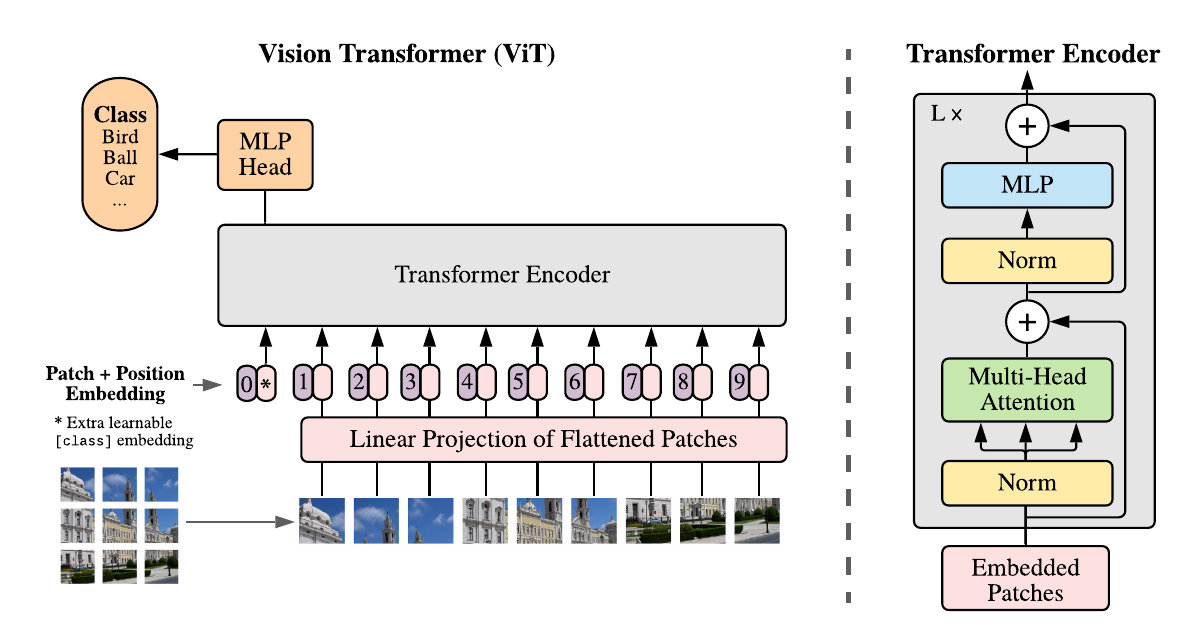}
\end{center}
   \caption{The architecture of Visual Transformer (ViT).}
\label{fig:vit}
\end{figure}

\textbf{iGPT}.
iGPT \cite{chen2020generative} is the first one to use the method of generating pre-training in the image field. It directly converts the image into a one-dimensional sequence as input to train a sequence Transformer to regress and predict pixels automatically. iGPT mainly includes pre-training and fine-tuning. The pre-training stage is especially by selecting two different pre-training objective functions, $L_{AR}$ or $L_{bert}$. The pre-training objective function of the autoregressive method is formulated as Eq. (\ref{eq:lar}).
\begin{equation}
L_{AR}=E_{x\sim X}[-log p(x)],
\label{eq:lar}
\end{equation}
where $p(x)=\prod\overline{n}_{i=1}p(x_{\pi_{i}}|x_{\pi_{1}},\ldots x_{\pi_{2}},\theta)$. The other one is the pre-training loss function of the Bert MLM method, and its expression is expressed as Eq. (\ref{eq:lbert}).
\begin{equation}
L_{bert}=E_{x\sim X}E_{M}\sum_{i\in M}[-log\, p(x_{i}|x_{[1,n]\setminus M})].
\label{eq:lbert}
\end{equation}

The iGPT model first preprocesses the image, ignoring the two-dimensional structure, and pulls it into a one-dimensional sequence. Then, $L_{AR}$ or $L_{bert}$ is selected as the pre-training objective function to perform autoregressive prediction of the next or mask pixel. Finally, the pre-trained model is used as a feature extractor, and then a linear classifier is used for classification prediction or through an end-to-end fine-tuning model on the target data for classification prediction. In this way, unsupervised generative training is performed so that the GPT-2 scale model can learn a strong image representation. It can also obtain excellent performance on image classification tasks.

\textbf{DeiT}.
Data-efficient image Transformers (DeiT) \cite{touvron2021training} was proposed by Facebook AI. A high-performance image classification model can be modeled with fewer data and fewer computing resources. DeiT mainly draws on the training strategy of CNN distillation and proposes a new distillation process based on a distillation token, which can copy the label estimated by the teacher network. This specific Transformer strategy is significantly better than the vanilla distillation method.
This particular Transformer distillation method adds a new distillation token to the original class token and patch token to interact with the self-attention layer together. The purpose of the distillation token is to replicate the hard label predicted by the teacher network. Both the class token and distillation token inputs are learned through back-propagation.

\textbf{CrossVit}.
Cross-attention multi-scale vision Transformer (CrossVit) \cite{chen2021crossvit} aims to learn the multi-scale feature representations in Transformer models for image classification. It is a two-branch Transformer that combines image blocks of different sizes to produce more robust image features. Besides, it comprises K multi-scale Transformer encoders, each of which is composed of an L-Branch and an S-Branch. The L-Branch uses coarse-grained patches to operate, and the S-Branch works on fine-grained patches. In contrast, L-Branch has more encoders and larger embedding dimensions. The above two branch output features are fused through Cross-Attention, and the CLS token is predicted using the two branches at the end. During the learning process, a learnable position embedding is used to learn position information for each token of the two branches.

\begin{table*}[]
\footnotesize
\centering
\caption{Performance Statistic of Transformer-based Methods for image classification on the Cifar 10, cifar 100, and ImageNet Data set. The methods, Pre-training data, image size, parameters, flops and the top-1 accuracy on the three data sets are shown in this table.}
\label{tab:classification}
\begin{tabular}{c|c|c|c|c|c|c|c}
\toprule
\textbf{Methods}             & \textbf{Pre-training data} & \textbf{image size} & \textbf{Paras(M)} & \textbf{Flops(G)} & \textbf{Cifar10(\%)} & \textbf{Cifar100(\%)} & \textbf{ImageNet(\%)} \\ \toprule
ViT-B/16 \cite{dosovitskiy2020image}          & ImageNet-1K      & 224                   & 86       & 743      & 98.13       & 87.13        & 77.91        \\ \hline
ViT-L/16 \cite{dosovitskiy2020image}           & ImageNet-21K     & 384                   & 307      & 5172     & 99.15       & 93.25        & 85.3         \\ \hline
ViT-L/16 \cite{dosovitskiy2020image}           & JFT-300M        & 384                   & 307      & 5172     & 99.42       & 93.9         & 87.76        \\ \hline
BoTNet-S1-59-T2 \cite{srinivas2021bottleneck}    & ImageNet-1K      & 224                   & 33.5     & 7.3      & -        & -         & 81.7         \\ \hline
BoTNet-S1-110-T4 \cite{srinivas2021bottleneck}   & ImageNet-1K      & 224                   & 54.7     & 10.9     & -        & -         & 82.8         \\ \hline
BoTNet-S1-128-T5 \cite{srinivas2021bottleneck}   & ImageNet-1K      & 256                   & 75.1     & 19.3     & -        & -         & 83.5         \\ \hline
DeiT-S \cite{touvron2021training}            & ImageNet-1K     & 224                   & 22.1     & 4.6      & 99.15       & 90.89        & 79.8         \\ \hline
DeiT-B \cite{touvron2021training}            & ImageNet-1K     & 224                   & 86.8     & 17.6     & 99.1        & 90.8         & 81.8         \\ \hline
DeiT-B$\uparrow$ \cite{touvron2021training}        & ImageNet-1k     & 224                   & 86       & 52.8     & 99.1        & 91.4         & 83.1         \\ \hline
CrossViT-S \cite{chen2021crossvit}        & ImageNet-1K     & 224                   & 26.7     & 5.6      & -        & -         & 81           \\ \hline
CrossViT-B \cite{chen2021crossvit}        & ImageNet-1K     & 224                   & 104.7    & 21.2     & -        & -         & 82.2         \\ \hline
CrossViT-15 \cite{chen2021crossvit}       & ImageNet-1K     & 224                   & 27.4     & 5.8      & 99          & 90.77        & 81.5         \\ \hline
CrossViT-18 \cite{chen2021crossvit}       & ImageNet-1K     & 224                   & 43.3     & 9.03     & 99.11       & 91.36        & 82.5         \\ \hline
CvT-13 \cite{wu2021cvt}            & ImageNet-1K     & 224                   & 20       & 4.5      & -        & -         & 81.6         \\ \hline
CvT-13$\uparrow$ \cite{wu2021cvt}           & ImageNet-1K     & 384                   & 20       & 16.3     & -        & -         & 83           \\ \hline
CvT-21 \cite{wu2021cvt}             & ImageNet-1K     & 224                   & 32       & 7.1      & -        & -         & 82.5         \\ \hline
CvT-21 \cite{wu2021cvt}             & ImageNet-1K     & 384                   & 32       & 24.9     & -        & -         & 83.3         \\ \hline
CvT-13 \cite{wu2021cvt}            & ImageNet-21K    & 384                   & 20       & 16       & 98.83       & 91.11        & 83.3         \\ \hline
CvT-21 \cite{wu2021cvt}             & ImageNet-21K    & 384                   & 32       & 24.9     & 99.16       & 92.88        & 84.9         \\ \hline
CvT-W24 \cite{wu2021cvt}            & ImageNet-21K    & 384                   & 277      & 193.2    & 99.39       & 94.09        & 87.7         \\ \hline
CSWin-T \cite{dong2021cswin}           & ImageNet-1K     & 384                   & 23       & 14       & -        & -         & 84.3         \\ \hline
CSWin-S \cite{dong2021cswin}            & ImageNet-1K     & 384                   & 35       & 22       & -        & -         & 85           \\ \hline
CSWin-B \cite{dong2021cswin}           & ImageNet-1K     & 384                   & 78       & 47       & -        & -         & 85.4         \\ \hline
CSWin-B \cite{dong2021cswin}           & ImageNet-21K    & 384                   & 78       & 47       & -        & -         & 87           \\ \hline
CSWin-L \cite{dong2021cswin}           & ImageNet-21K    & 384                   & 173      & 96.8     & -        & -         & 87.5         \\ \hline
DeepVit-L \cite{zhou2021deepvit}         & -              & -                  & 55       & 12.5     & -        & -         & 82.2         \\ \hline
DeepVit-S \cite{zhou2021deepvit}         & -              & -                  & 27       & 6.2      & -        & -         & 81.4         \\ \hline
ViL-S \cite{zhang2021multi}             & -              & 224                   & 24.6     & 4.9      & -        & -         & 82.4         \\ \hline
ViL-M \cite{zhang2021multi}             & -              & 224                   & 39.7     & 8.7      & -        & -         & 83.5         \\ \hline
ViL-B \cite{zhang2021multi}             & -   & 224                   & 55.7     & 13.4     & -        & -         & 83.7         \\ \hline
ViL-LS-M \cite{zhang2021multi}          & -   & 224                   & 39.8     & 8.7      & -        & -         & 83.8         \\ \hline
ViL-LS-M \cite{zhang2021multi}          & -   & 384                   & 39.9     & 28.7     & -        & -         & 84.4         \\ \hline
ViL-LS-B \cite{zhang2021multi}          & -   & 224                   & 55.8     & 13.4     & -        & -         & 84.1         \\ \hline
ViL-B \cite{zhang2021multi}             & ImageNet-21K    & 384                   & 56       & 43.7     & -        & -         & 86.2         \\ \hline
PVT-S \cite{wang2021pyramid}           & -   & -         & 24.5     & 3.8      & -        & -         & 79.8         \\ \hline
PVT-M \cite{wang2021pyramid}             & -   & -         & 44.2     & 6.7      & -        & -         & 81.2         \\ \hline
PVT-L \cite{wang2021pyramid}             & -   & -         & 61.4     & 9.8      & -        & -         & 81.7         \\ \hline
Swin-B \cite{liu2021swin}            & ImageNet-21K    & 224                   & 88       & 15.4     & -        & -         & 85.2         \\ \hline
Swin-L \cite{liu2021swin}            & ImageNet-21K    & 384                   & 88       & 47       & -        & -         & 86.4         \\ \hline
Swin-L \cite{liu2021swin}            & ImageNet-1K      & 384                   & 197      & 103.9    & -        & -         & 87.3         \\ \hline
TNT-S \cite{han2021Transformer}            & ImageNet-1K      & 224                   & 23.8     & 5.2      & -        & -         & 81.3         \\ \hline
TNT-B \cite{han2021Transformer}             & ImageNet-1K      & 384                   & 65.6     & 14.1     & 99.1        & 91.1         & 83.9         \\ \hline
TNT-S \cite{han2021Transformer}             & ImageNet-1K      & 384                   & 23.8     & -     & 98.7        & 90.1         & 83.1         \\ \hline
Visformer-S \cite{chen2021visformer}       & -   & -         & 40.2     & 4.9      & -        & -         & 82.19        \\ \hline
Visformer-Ti \cite{chen2021visformer}      & -   & -         & 10.3     & 1.3      & -        & -         & 78.6         \\ \hline
PSViT-2D-Base \cite{chen2021psvit}     & -   & -         & -     & 15.5     & -        & -         & 82.9         \\ \hline
PSViT-1D-Base \cite{chen2021psvit}      & -   & -         & -     & 18.9     & -        & -         & 82.6         \\ \hline
PSViT-1D-Small \cite{chen2021psvit}    & -   & -         & -     & 4.9      & -        & -         & 80.7         \\ \hline
PSViT-2D-Small \cite{chen2021psvit}    & -   & -         & -     & 4.4      & -        & -         & 81.6         \\ \hline
iGPT \cite{chen2020generative}              & ImageNet-1K      & 224                   & 1362     & -     & 99          & 88.5         & -         \\ \hline
T2T-ViT -19 \cite{yuan2021tokens}        & ImageNet-1K      & 224                   & 39.2     & 8.9      & 98.3        & 89           & 81.9         \\ \hline
T2T-ViT -14 \cite{yuan2021tokens}        & ImageNet-1K      & 224                   & 21.5     & 5.2      & 97.5        & 88.4         & 81.5         \\ \hline
Nested-Transformer \cite{zhang2021aggregating} & ImageNet        & 224                   & 90.1     & -     & 97.2        & 82.56        & 83.8         \\ \hline
CPVT-S \cite{chu2021conditional}             & ImageNet-1K      & 224                   & 23       & 4.6      & -        & -         & 80.5         \\ \hline
CPVT-S-GAP \cite{chu2021conditional}        & ImageNet-1K      & 224                   & 23       & 4.6      & -        & -         & 81.5         \\ \hline
CPVT-B \cite{chu2021conditional}            & ImageNet-1K      & 224                   & 88       & 17.6     & -        & -         & 82.3         \\ \hline
Twins-SVT-S \cite{chu2021twins}       & ImageNet-1K      & 224                   & 24       & 2.9      & -        & -         & 81.7         \\ \hline
Twins-SVT-B \cite{chu2021twins}       & ImageNet-1K      & 224                   & 56       & 8.6      & -        & -         & 83.2         \\ \hline
Twins-SVT-L \cite{chu2021twins}       & ImageNet-1K      & 224                   & 99.2     & 15.1     & -        & -         & 83.7         \\ \hline
Shuffle-T \cite{chen2021visformer}        & ImageNet-1K      & 224                   & 29       & 4.6      & -        & -         & 82.5         \\ \hline
Shuffle-S \cite{chen2021visformer}          & ImageNet-1K      & 224                   & 50       & 8.9      & -        & -         & 83.5         \\ \hline
Shuffle-B \cite{chen2021visformer}         & ImageNet-1K      & 224                   & 88       & 15.6     & -        & -         & 84           \\ \hline
VOLO-D1 \cite{yuan2021volo}          & ImageNet-1K      & 224                   & 27       & 6.8      & -        & -         & 84.2         \\ \hline
VOLO-D2 \cite{yuan2021volo}           & ImageNet-1K      & 224                   & 59       & 14.1     & -        & -         & 85.2         \\ \hline
VOLO-D3 \cite{yuan2021volo}           & ImageNet-1K      & 224                   & 86       & 20.6     & -        & -         & 85.4         \\ \hline
VOLO-D4 \cite{yuan2021volo}           & ImageNet-1K      & 224                   & 193      & 43.8     & -        & -         & 85.7         \\ \hline
VOLO-D5 \cite{yuan2021volo}           & ImageNet-1K      & 224                   & 296      & 69       & -        & -         & 86.1         \\ \hline
\end{tabular}
\end{table*}


\subsubsection{Remote sensing image classification}
\textbf{SST}.
Spatial-Spectral Transformer \cite{he2021spatial} is proposed for hyperspectral image classification. In the SST model, CNN extracts the spatial features, and the improved Transformer can capture the sequential spectra relationships. The dynamic feature augmentation (FA) is proposed to relieve the overfitting and improve the model's generalization.
Besides, transfer learning is combined with the SST (T-SST-L), which addresses the issue of limited training samples in HSI classification. The proposed SST model with different strategies shows competitive performance on the hyperspectral datasets Salinas, Pavia, and Indian Pines.

\textbf{SSTN}.
SSTN \cite{zhong2021spectral} is a novel spectral-spatial Transformer network, which can overcome the constraints of convolution kernels. It mainly contains two modules, including spatial attention and spectral association modules. A factorized architecture search (FAS) framework is designed to determine the layer-level operation choices and block-level orders of SSTN. The proposed model is evaluated on the HSI benchmarks, including the Indian Pines, the Kennedy Space Center, the University of Houston, and the Pavia Center (PC) datasets. SSTN shows the model's effectiveness and proves the FAS strategy.

\subsubsection{Medical image classification}
\textbf{GasHis-Transformer}.
GasHis-Transformer \cite{chen2021gashis} is a multi-scale visual Transformer model, which is proposed for Gastric Histopathological Image Classification (GHIC).
It mainly consists of a global information module and a local information module to extract histopathological features effectively.
The model has been evaluated on a public hematoxylin and eosin (H\&E) stained gastric histopathological data set, immunohistochemically stained images on a lymphoma image, and a breast cancer dataset. GasHis-Transformer shows high classification performance and great potential in GHIC tasks.

\subsubsection{Remarks}
Transformer is widely used in various tasks of NLP at first. Subsequently, many derivative models appeared, such as BERT \cite{devlin2019bert}, GPT \cite{radford2018improving}, GPT-2 \cite{radford2019language} and other models. Until 2020, ViT is proposed to apply Transformer to image classification tasks in CV. It shows a top-1 accuracy of 85.15\% on ImageNet, surpassing the accuracy of most CNNs at the time on ImageNet. Therefore, it has attracted the attention of researchers. Subsequently, there are more and more relevant models for Transformer-based image classification tasks, such as Swin Transformer, BoTNet, etc.

The Transformer is applied to natural image classification, remote sensing image classification, and medical image classification in the classification task. In terms of model structure, the Transformer model for image classification began to turn to the combination of CNN and Transformer, which again surpassed the pure Transformer model in performance. In recent years, the performance statistics of related image classification Transformer models on ImageNet, Cifar10, and Cifar100 are shown in TABLE \ref{tab:classification}. The three data sets are all counted: the statistical model, pre-training data set, image size, model parameter amount, and calculation amount and top-1 accuracy. JFT data set \cite{sun2017revisiting} and ImageNet data set \cite{deng2009imagenet} are the main pre-training data sets. These experimental results statistics can provide a corresponding reference for some researchers.

Transformer has relatively much-related research on natural data sets in image classification applications. The related models of remote sensing images and medical images are less developed. Transferring the Transformer model used in the natural data sets to the remote sensing images or medical image classification will become the research direction. Besides, Transformer for few-shot classification is also a research direction. For example, the universal representation Transformer (URT) is proposed in \cite{liu2020universal}, which learns the universal features for few-shot classification.


\subsection{Object detection}
\begin{figure}[htp]
\setlength{\belowcaptionskip}{-0.1cm}
   \includegraphics[width=1.0\linewidth]{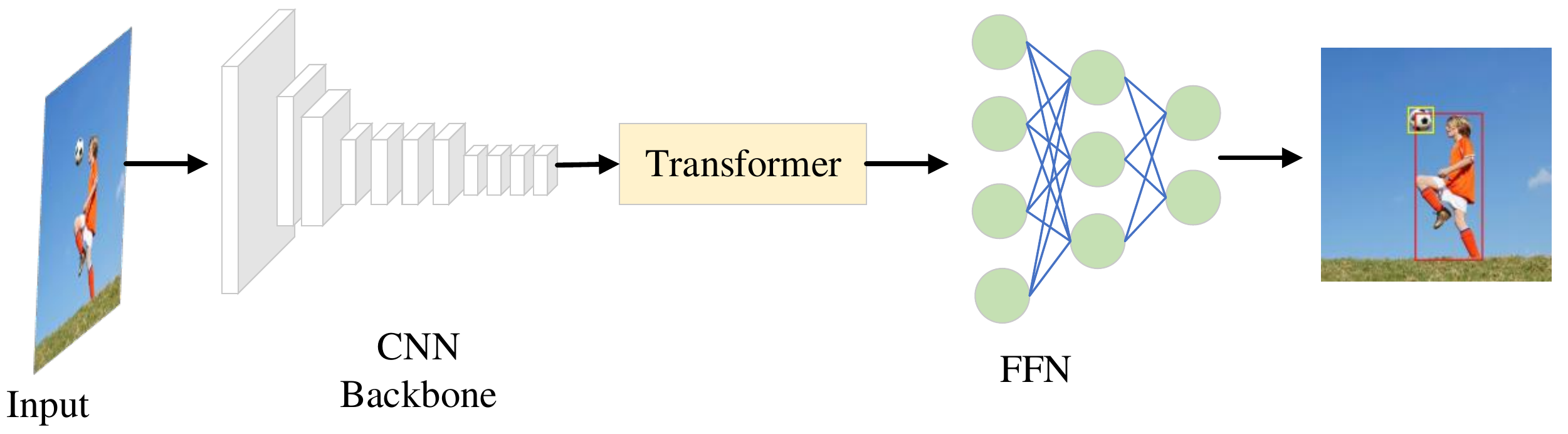}
   \caption{Basic structure of end-to-end detectors based on Transformer.}
\label{fig:detector}
\end{figure}

The overall framework of Transformer-based image detection is shown in Fig. \ref{fig:detector}. After the input image CNN backbone network performs feature extraction, it is encoded and decoded by Transformer, and then the object category and its bounding box in the image are predicted by the feed forward network. There are three components in the current target detection algorithm based on deep learning, including backbone, neck, and head. The backbone network extracts features, the neck extracts some more complex features, and the head calculates the predicted output. The Transformer-based methods for 2D and 3D object detection are mainly introduced in this subsection.
\subsubsection{2D object detection}

\textbf{DETR}.
DEtection Transformer (DETR)\cite{carion2020end} is an end-to-end target detection method based on Transformer. It mainly consists of CNN backbone network, Transformer encoder-decoder structure, and feed-forward network (FFN). First, CNN backbone is adopted to extract the feature of the input image. And then, the extracted feature is converted into a one-dimensional feature map and sent to the Transformer encoder. Subsequently, the Transformer decoder uses itself and the encoder-decoder attention mechanism to decode these embeddings into the bounding box coordinates. Finally, the feed-forward neural network predicts the normalized center coordinates, height, and width of the bounding box, while the linear layer uses the softmax function to predict the category label.

\textbf{Deformable DETR}.
The Transformer attention module has certain limitations when processing feature maps, the network convergence speed is slow, and the feature spatial resolution is limited. Unlike DETR, the core of variable DETR is that its attention module only focuses on a set of crucial sampling points around the reference point. It also has obvious advantages in the detection of small objects.
Deformable DETR \cite{zhu2020deformable} combines the advantages of sparse spatial sampling of deformable convolution with the relational modeling capabilities of Transformers. It mainly uses deformable attention to replace the original Transformer attention.

\textbf{UP-DETR}.
UP-DETR \cite{dai2021up}  is an unsupervised pre-training Transformer for target detection. For a given image, the UP-DETR performs random crop patches and then provides them as queries to the decoder. UP-DETR is pre-trained and can detect these query patches from the original image. It mainly includes pre-training stage and fine-tuning stage. It mainly focuses on the unsupervised and training phases. In the UP-DETR model, the input image is first sent to CNN to extract its feature. And then, the extracted feature map and the position codes are sent to the multi-layer Transformer's encoder in DETR. Random cropped query patch extracts features through CNN with global average pooling and then is flatted as the target query object and sent to the Transformer encoder. The encoder can predict the bounding box corresponding to the randomly cut query patch.

\textbf{ACT}.
DETR requires amount of computing resources for training and inference due to the high-resolution input. To effectively relief this problem, the adaptive clustering Transformer (ACT) \cite{carion2020end} is proposed. It uses locality-sensitive hashing to cluster query features adaptively and uses prototype key interaction to approximate query key interaction, which can directly replace the original self-attention module. As a whole, the computing resources are reduced, and a certain model accuracy can be guaranteed.

\textbf{RelationNet++}.
Different detectors express the target in different ways. 2-stage detectors usually adopt ROI features, most anchor-free methods commonly adopt point features, and RepPoints methods usually utilize point features sets. The enhanced ATSS detector, named RelationNet++ is proposed in \cite{chi2020relationnet++}. It mainly uses the attention mechanism to use different types of feature expressions to enhance the features of a detector to detect the target and make full use of the advantages of multiple types of features (ROI/center/corner). The proposed bridging visual representations (BVR) module includes key sampling and shared location embedding. This module combines multiple expressions into popular detection frameworks to improve the performance of each detector. RelationNet++‘*’ means the multi-scale testing of RelationNet++ version.

\textbf{TSP-FCOS and TSP-RCNN}.
To alleviate the slow convergence problem caused by the instability of the bipartite graph used by the DETR decoder and Hungary loss, TSP-FCOS and TSP-RCNN are proposed in \cite{sun2021rethinking}. They both cut the decoder part. And simultaneously, the self-attention of the encoder is only performed on selected examples. It applies the latest feature of interest (FoI) module to TSP-FCOS to help Transformer encode multi-scale features. At the same time, a new bipartite graph matching was developed for these two models to accelerate convergence during training.

\textbf{YOLOS}.
YOLOS \cite{fang2021you} proposes to perform 2D target recognition on the Transformer from a pure sequence to sequence perspective with a minimal 2D spatial structure. The thesis is based on the original ViT architecture and appropriately adjusted on the detection model DETR. YOLOS can easily adapt to different Transformer structures and perform arbitrary-dimensional target detection without the need for precise spatial structure or geometric structure. At the same time, YOLOS uses DET as the agent of target expression to avoid the inductive bias caused by the prior knowledge of 2D structure and task and reveal the characteristics of Transformer in target detection as unbiased as possible.

\textbf{Conditional DETR}.
To alleviate the slow convergence speed of DETR, the conditional cross-attention mechanism for fast DETR training, named conditional DETR \cite{meng2021conditional} is proposed. It learns the conditional space query from the decoder embedding, which is used for multi-head cross attention of the decoder. Each cross-attention head can focus on bands containing different regions, effectively reducing the spatial range of different regions for positioning object classification and frame regression, thereby relaxing the dependence on content embedding and simplifying training. Experiments show that convergence speed on various backbone networks (R50, R101, DC5-R50, DC5-R101) is about 6.7-10 times faster than DETR.

\textbf{SMCA}.
A spatial modulation cooperative attention (SMCA) \cite{gao2021fast} mechanism is proposed to speed up the convergence of DETR. The core idea is to regress and perceive common attention in DETR by limiting the response to a higher position near the initially estimated bounding box. It replaces the original common attention mechanism in DETR with the proposed SMCA, which can effectively improve the convergence speed of the model. The experiment verifies the effectiveness of the proposed model on the COCO data set. SMCA with multi-scale features (MS) achieved an accuracy of 45.6\% mAP on 2017 COCO VAL.
\begin{table*}[htp]
\small
\centering
\caption{Comparison Transformer-based Methods for Object Detection on COCO 2017 VAL Set. ``Multi-scale" means the multi-scale testing. ``TTA" indicates test-time augmentations including horizontal flip and multi-scale testing.}
\label{tab:objectdetection}
\begin{tabular}{c|c|c|c|c|c|c|c|c|c|c}
\toprule
\textbf{Methods}         & \textbf{Backbone}              & \textbf{Paras(M)} & \textbf{GFlops(G)} & \textbf{fps}  & \textbf{AP}  & \textbf{AP50} & \textbf{AP75} & $\textbf{AP}_{s}$  & $\textbf{AP}_{m}$  & $\textbf{AP}_{l}$  \\ \toprule
PSViT-2D-Tiny \cite{chen2021psvit}                    & -                  & -     & 1.3       & - & 40.8 & 64.7 & 44   & 25.3 & 43.8 & 53.9 \\ \hline
RelationNet++ \cite{chi2020relationnet++}                  & ResNeXt-64x4d-101-DCN & -     & -      & - & 50.3 & 69   & 55   & 32.8 & 55   & 65.8 \\ \hline
RelationNet++ * \cite{chi2020relationnet++}                 & ResNeXt-64x4d-101-DCN & -     & -      & - & 52.7 & 70.4 & 58.3 & 35.8 & 55.3 & 64.7 \\ \hline
\multirow{4}{*}{DETR \cite{carion2020end}}            & ResNet-50             & 41       & 86        & 28   & 42   & 62.4 & 44.2 & 20.5 & 45.8 & 61.1 \\ \cline{2-11}
                                 & DC5                   & 41       & 187       & 12   & 43.3 & 63.1 & 45.9 & 22.5 & 47.3 & 61.1 \\ \cline{2-11}
                                 & R101                  & 60       & 152       & 20   & 43.5 & 63.8 & 46.4 & 21.9 & 48   & 61.8 \\ \cline{2-11}
                                 & DC5-R101              & 60       & 153       & 10   & 44.9 & 64.7 & 47.7 & 23.7 & 49.5 & 62.3 \\ \hline
\multirow{5}{*}{Deformable DETR \cite{zhu2020deformable}} & ResNet-50             & -     & -      & - & 46.9 & 66.4 & 50.8 & 27.7 & 49.7 & 59.9 \\ \cline{2-11}
                                 & ResNet-101            & -     & -      & - & 48.7 & 68.1 & 52.9 & 29.1 & 51.5 & 62   \\ \cline{2-11}
                                 & ResNeXt-101           & -     & -      & - & 49   & 68.5 & 53.2 & 29.7 & 51.7 & 62.8 \\ \cline{2-11}
                                 & ResNeXt-101-DCN       & -     & -      & - & 50.1 & 69.7 & 54.6 & 30.6 & 52.8 & 64.7 \\ \cline{2-11}
                                 & ResNeXt-101-DCN(TTA)  & -     & -      & - & 52.3 & 71.9 & 58.1 & 34.4 & 54.4 & 65.6 \\ \hline
ACT(L=32) \cite{zheng2020end}                        & ResNet-50             & -     & 168.9     & 16   & 42.6 & - & - & 22.5 & 46.8 & 61.1 \\ \hline
ACT+MKTD(L=32) \cite{zheng2020end}                   & ResNet-50             & -     & 168.9     & 16   & 43.1 & - & - & 22.2 & 47.1 & 61.4 \\ \hline
UP-DETR \cite{dai2021up}                          & ResNet-50             & 41       & 86        & 28   & 40.5 & 60.8 & 42.6 & 19   & 44.4 & 60   \\ \hline
UP-DETR+ \cite{dai2021up}                        & ResNet-50             & 41       & 86        & 28   & 42.8 & 63   & 45.3 & 20.8 & 47.1 & 61.7 \\ \hline
TSP-FCOS \cite{sun2021rethinking}                         & ResNet-50             & 51.5     & 189       & 15   & 43.1 & 62.3 & 47   & 26.6 & 46.8 & 55.9 \\ \hline
TSP-RCNN \cite{sun2021rethinking}                        & ResNet-50             & 64       & 188       & 11   & 43.8 & 63.3 & 48.3 & 28.6 & 46.9 & 55.7 \\ \hline
TSP-RCNN+ \cite{sun2021rethinking}                       & ResNet-50             & 64       & 188       & 11   & 45   & 64.5 & 49.6 & 29.7 & 47.7 & 58   \\ \hline
\multirow{4}{*}{Pix2Seq \cite{chen2021pix2seq}}         & R50                   & 37       & -      & - & 43   & 61   & 45.6 & 25.1 & 46.9 & 59.4 \\ \cline{2-11}
                                 & R101                  & 56       & -      & - & 44.5 & 62.8 & 47.5 & 26   & 48.2 & 60.3 \\ \cline{2-11}
                                 & R50-DC5               & 38       & -      & - & 43.2 & 61   & 46.1 & 26.6 & 47   & 58.6 \\ \cline{2-11}
                                 & R101-DC5              & 57       & -      & - & 45   & 63.2 & 48.6 & 28.2 & 48.9 & 60.4 \\ \hline
YOLOS-S \cite{fang2021you}                          & DeiT-S                & 30.7     & 200       & 7    & 36.1 & 56.4 & 37.1 & 15.3 & 38.5 & 56.1 \\ \hline
YOLOS-S(MS)\cite{fang2021you}             & DeiT-S                & 27.9     & 179       & 5    & 37.6 & 57.6 & 39.2 & 15.9 & 40.2 & 57.3 \\ \hline
YOLOS-B \cite{fang2021you}                         & DeiT-B                & 127      & 537       & - & 42   & 62.2 & 44.5 & 19.5 & 45.3 & 62.1 \\ \hline
Efficient DETR \cite{yao2021efficient}                   & ResNet-50             & 32       & 159       & - & 44.2 & 62.2 & 48   & 28.4 & 47.5 & 56.6 \\ \hline
Efficient DETR* \cite{yao2021efficient}                 & ResNet-50             & 35       & 210       & - & 45.1 & 63.1 & 49.1 & 28.3 & 48.4 & 59   \\ \hline
SMCA \cite{gao2021fast}                            & ResNet-50             & 40       & 152       & 10   & 43.7 & 63.6 & 47.2 & 24.2 & 47   & 60.4 \\ \hline
SMCA(MS) \cite{gao2021fast}                           & ResNet-50             & 40       & 152       & 10   & 45.6 & 65.5 & 49.1 & 25.9 & 49.3 & 62.6 \\ \hline
Conditional DETR \cite{meng2021conditional}                & ResNet-50             & 44       & 90        & - & 43   & 64   & 45.7 & 22.7 & 46.7 & 61.5 \\ \hline
Conditional DETR-DC5 \cite{meng2021conditional}            & ResNet-50             & 44       & 195       & - & 45.1 & 65.4 & 48.5 & 25.3 & 49   & 62.2 \\ \hline
ViT-B/16*-FRCNN \cite{beal2020toward}                  & -                  & -     & -      & - & 37.8 & 57.4 & 40.1 & 17.8 & 41.4 & 57.3 \\ \hline
ViT-B/16-FRCNN \cite{beal2020toward}                   & -                  & -     & -      & - & 36.6 & 56.3 & 39.3 & 17.4 & 40   & 55.5 \\ \hline
\end{tabular}
\end{table*}

\subsubsection{3D object detection}
3D target detection is one of the essential contents of autonomous driving. In the process of automatic driving, accurately locating the obstacle's position is more important than identifying the type of obstacle. It can be conducive to proper planning of routes and realize more intelligent and safe automatic driving functions.
3D object detectors can be divided into single-mode (Lidar) and multi-mode (Lidar or Camera) according to the input type. Besides, it is divided into Point Clouds, Voxel, Graph, and 2D View according to feature extraction.
In addition to the 3D CNN target detector, the current 3D detector based on Transformer has also been continuously developed. Here, we sort out several emerging 3D Transformer target detectors.

\textbf{TCTR}.
Each original point cloud data video frame is first converted into a two-dimensional pseudo image frame. Then, the proposed time channel Transformer, TCTR \cite{yuan2021temporal} module, is used to generate a representation containing time channel information, and then feature refinement and detection heads are performed to generate detection results.

\textbf{Pointformer}.
Pointformer \cite{pan20213d} is a Transformer backbone network specially designed for 3D point clouds, which can effectively learn 3D point cloud data characteristics. A Pointformer block comprises the local, the local-global, and the global Transformer. The Local Transformer module is used to model the interaction between points in the local area to learn context-related regional features at the object level. The global Transformer aims to understand context-aware representations at the scene level. The local-global Transformer integrates local features with high-resolution global features to further capture the dependencies between multi-scale representations. Thus, Pointformer can effectively combine high-resolution and low-resolution elements, use multi-scale cross-attention to fuse high-resolution and low-resolution features, and perceive contextual representation.

\textbf{CT3D}.
CT3D \cite{sheng2021improving} is proposed to alleviate the limitation of the limited ability of the previous methods in terms of the rich context correlation between the capture points. It mainly uses a high-quality regional proposal network and channel-wise Transformer architecture, uses the proposal's key points for spatial context modeling, and learns attention propagation in the coding module to map the proposal to the point embedding. Subsequently, a new channel-based decoding module enriches query key interactions through channel-based re-weighting, effectively merging multi-level contexts. The experiment has superior performance on the KITTI test 3D detection benchmark and has certain scalability.

\textbf{3DETR}.
3DETR \cite{misra2021end} is also an end-to-end 3D point cloud target detection model based on Transformer. The 3D point cloud data is input to generate a set of points through down-sampling. The Transformer encoder uses multiple layers of self-attention to generate features for each point. The generated point features and a set of queries are embedded in the decoder, and then the 3D bounding box is output. 3DETR found that the standard Transformer with non-parametric query and Fourier position embedding is competitive with a dedicated architecture that utilizes a 3D specific operator library with manually adjusted hyperparameters. 3DETR outperforms the complete and highly optimized VoteNet baseline by 9.5\% on the challenging ScanNetV2 data set.

\textbf{VoTr}.
Voxel Transformer (VoTr) \cite{mao2021voxel} is also a voxel-based Transformer network for 3D point cloud target detection. In the paper, the remote association between voxels is realized through the self-attention of Transformer, which effectively alleviates the limitations of the previous 3D point cloud detector. At the same time, the paper also proposes sparse voxel modules and sub-manifold voxel modules, which can effectively operate on empty and non-empty pixels. The proposed local attention and expanded attention can further expand the range of attention and speed up the query process of multi-head attention by fast voxel query. It contains a series of sparse and sub-manifold voxel modules and can be applied to most voxel-based detectors.

\textbf{T3D}.
Aiming at the results of the existing 3D target detection based on voting methods that are far away from the center of the real object due to inaccurate voting. Transformer3D-Det (T3D) \cite{zhao2021Transformer3d} introduces a Transformer-based voting refinement module to improve the voting results of VoteNet. It is composed of three parts, including vote generation module, vote refinement module and bounding box generation module. First, the voting generation model is used to generate multiple coarse vote clusters on the input point cloud. Secondly, the proposed Transformer-based voting refinement module performs further detailed voting on the clustered coarse votes. Finally, the bounding box generation module takes the refined voting clusters as input and generates the final detection result of the input point cloud. In the training process, T3D proposes to use a new non-voting loss function to constrain, which can effectively improve the voting accuracy.

\subsubsection{Remarks}
Because the Transformer method has an infinite receptive field, the introduction of Transformer can save a lot of manual design operations in target detection, such as NMS, region proposal, and so on. DETR is the pioneering work of Transformer applied to target detection. YOLOS is a series of ViT-based target detection models with as few modifications and inductive biases as possible. Anchor DETR draws on the Anchor Point mechanism in the CNN target detector, so that each query is based on a specific Anchor Point. Anchor DETR obtains better performance and faster running speed than DETR.
In addition, there are many related variants of DETR. Aiming at the problem of the slow convergence speed of DETR, the researchers proposed TSP-FCOS \& TSP-RCNN. Deformable DETR uses deformable convolution, which effectively focuses on sparse space positioning, to alleviate the two problems of slow DETR convergence and low detection accuracy of small targets. ACT mainly alleviates the redundancy of the attention map in DETR and the problem of feature redundancy as the encoder deepens.

This paper counts some Transformer's target detection models, backbone networks, parameters, calculations, and performance on COCO 2017val, as shown in TABLE \ref{tab:objectdetection}. This part aims to provide references for researchers who study target detection. The Transformer-based target detection model has made good progress in performance, but still needs to be improved in terms of real-time performance, which may become a major research trend. Besides, the paper \cite{Chai_2021_CVPR}  proposes to learn 3D representation directly from 2D perspective range image view. It designs a 2D convolution network architecture, which can carry the 3D spherical coordinates of each pixel in the whole network. Besides, it owns higher performance, higher efficiency, and less parameters. This provides a researching direction to extend 2D CNNs or Transformer to 3D representation application.

\subsection{Image Segmentation}
\begin{figure}[htp]
\setlength{\belowcaptionskip}{-0.1cm}
   \includegraphics[width=1.0\linewidth]{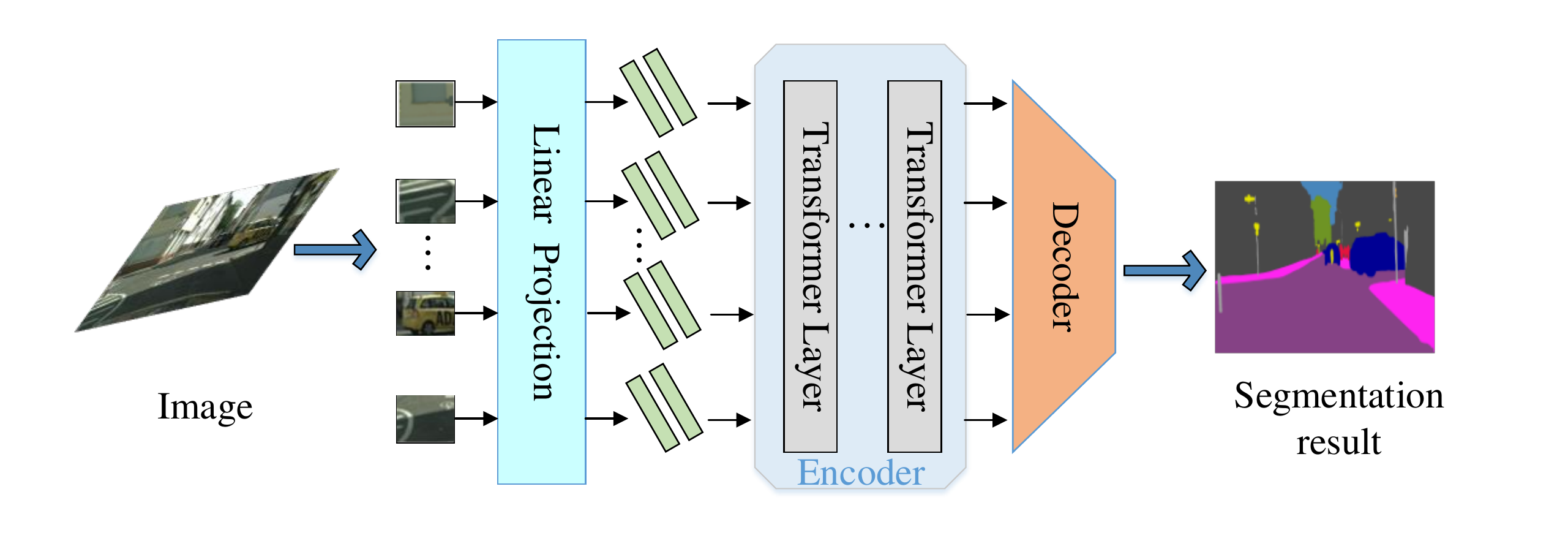}
   \caption{The basic structure of pure Transformer-based methods for image segmentation.}
\label{fig:segmentor}
\end{figure}

The example of a pure Transformer encoding-decoding architecture for image segmentation is shown as Fig. \ref{fig:segmentor}. According to statistics, the existing Transformer-based image segmentation methods are mainly used in medical image segmentation, point cloud segmentation, remote sensing image segmentation, and so on \cite{li2017partitioned,liang2020polytransform}. Among them, there are relatively many researches on medical image segmentation, such as CoTr \cite{xie2021cotr}, UNETR \cite{hatamizadeh2021unetr} and so on.
\subsubsection{Natural image Segmentation}
\textbf{Segmenter}.
As a pure Transformer encoding-decoding architecture, Segmenter \cite{strudel2021segmenter} utilizes the global image context of the model's each layer. Based on ViT, Segmenter divides the image into patches, maps them into a linear embedding sequence, and encodes them with an encoder. Then the mask Transformer decodes the output of the encoder and class embedding. The argmax is applied to classify each pixel after upsampling, and the final pixel segmentation map is output. The decoding stage adopts a simple method of jointly processing image blocks and class embedding. The mask Transformer decoder can directly perform panoramic segmentation by replacing class embedding with object embedding.
\begin{table*}[htp]
\centering
\caption{Comparison performance of Transformer-based Methods on Synapse multi-organ CT data set. DSC means average dice score
\% and HD represents average hausdorff distance in mm, and dice score \% for different organ.}
\label{tab:ct}
\small
\begin{tabular}{c|c|c|c|c|c|c|c|c|c|c}
\toprule
\textbf{Methods}         & \textbf{DSC$\uparrow$}  & \textbf{HD$\downarrow$}                       &\textbf{Aorta} & \textbf{Gallbladder} & \textbf{Kidney(L)} & \textbf{Kidney(R)} & \textbf{Liver} & \textbf{Pancreas} & \textbf{Spleen} & \textbf{Stomach} \\ \toprule
R50\_ViT \cite{dosovitskiy2020image}        & 71.29 & 32.87                     & 73.73 & 55.13       & 75.8      & 72.2      & 91.51 & 45.99    & 81.99  & 73.95   \\ \hline
TransUnet \cite{chen2021transunet}     & 77.48 & 31.69                     & 87.23 & 63.13       & 81.87     & 77.02     & 94.08 & 55.86    & 85.08  & 75.62   \\ \hline
SwinUnet \cite{cao2021swin}       & 79.13 & 21.55                     & 85.47 & 66.53       & 83.28     & 79.61     & 94.29 & 56.58    & 90.66  & 76.6    \\ \hline
LeVit-UNet-128s \cite{xu2021levit}  & 73.69 & 23.92                     & 86.45 & 66.13       & 79.32     & 73.56     & 91.85 & 49.25    & 79.29  & 63.7    \\ \hline
LeVit-UNet-192 \cite{xu2021levit} & 74.67 & 18.86                     & 85.69 & 57.37       & 79.08     & 75.9      & 92.05 & 53.53    & 83.11  & 70.61   \\ \hline
LeVit-Unet-384 \cite{xu2021levit}  & 78.53 & 16.84                     & 87.33 & 62.23       & 84.61     & 80.25     & 93.11 & 59.07    & 88.86  & 72.76   \\ \hline
TransClaw U-Net \cite{chang2021transclaw} & 78.09 & 26.38                     & 85.87 & 61.38       & 84.83     & 79.36     & 94.28 & 57.65    & 87.74  & 73.55   \\ \hline
LiteTrans \cite{xu2021litetrans}    & 77.91 & 29.01                     & 85.87 & 62.22       & 83.21     & 77.1      & 94.45 & 57.6     & 86.52  & 76.3    \\ \hline
ViTBIS \cite{sagar2021vitbis}       & 80.45 & 21.24                     & 86.41 & 66.8        & 83.59     & 80.12     & 94.56 & 56.9     & 91.28  & 76.82   \\ \hline
nnFormer \cite{zhou2021nnformer}        & \textbf{87.4}  & \multicolumn{1}{l|}{-} & \textbf{92.04} & \textbf{71.09}       & \textbf{87.64}     & \textbf{87.34}     & \textbf{96.53} & \textbf{82.49}   & \textbf{92.91}  & \textbf{89.17}   \\ \hline
AFTer-Unet \cite{yan2021after}     & 81.02 & \multicolumn{1}{l|}{-} & 90.91 & 64.81       & 87.9      & 85.3      & 92.2  & 63.54    & 90.99  & 72.48   \\ \hline
\end{tabular}
\end{table*}

\textbf{Maskformer}.
Mask classification is general enough in mask classification to effectively achieve semantic and instance-level segmentation. Based on this, Maskformer \cite{cheng2021per} is proposed to associate each mask with a single global class label prediction, which can effectively simplify scene and panorama segmentation tasks. Maskformer achieves 55.6\% mIoU accuracy on ADE20K and 52.7\% PQ accuracy on COCO.

\textbf{UperNet}.
Unified perceptual parsing, namely UPP, attempts to parse the multi-level visual concept of an image at once. When FCN solves the UPP problem, the depth convolution design structure is complex, the down-sampling rate is fast, and the defect that deeper feature maps are not conducive to segmenting low-level features of images. To improve the defects above, a multi-task framework, UperNet \cite{xiao2018unified} is proposed. It is based on Feature Pyramid Network (FPN) and uses features from multiple semantic levels, making the model more suitable for scene classification. It can learn from heterogeneous image annotations and is able to efficiently segment a wide range of concepts from images.

\textbf{Segformer}.
SegFormer \cite{xie2021segformer} unifies Transformer and lightweight multi-layer perceptron encoder. Layered Transformer encoder and avoiding complicated decoder are two advantages of SegFormer. The proposed MLP decoder aggregates information of different layers and can provide a powerful characterization ability combining local and global attention. SegFormer-B4 achieved 50.3\% mIoU in ADE20K.

\textbf{Max-DeepLab}.
MaX-DeepLab \cite{wang2021max} is an end-to-end model for panoramic segmentation based on Transformer, which simplifies the current pipeline that heavily relies on agent subtasks and manual design components. MaX-DeepLab directly uses the mask of the dual-path architecture to predict the mask of the class marker by the Transformer. The input of the dual-path Transformer is an image and a global memory respectively. MaX-DeepLab achieved 51.3\% PQ on the COCO test development set.

\subsubsection{Medical image Segmentation}
The CNN segmentation model U-Net has become a standard medical image segmentation network and achieved great success. At present, most of the image segmentation networks based on Transformer are also combined with U-Net. It mainly uses the innate global self-attention mechanism to solve the problem that U-Net usually shows the inherent locality of convolution operation is modeling remote dependencies.

\textbf{CoTr}.
Transformer can effectively solve the problem that CNN only has small receptive field and can not capture long-distance dependence. CoTr \cite{xie2021cotr} is a combination framework based on CNN and Transformer for accurate 3D medical image segmentation. It can effectively use the characteristics of Transformer and alleviate the problems of long training time and high computational overhead caused by high resolution. CoTr is mainly composed of CNN encoder, detrans encoder and decoder. It mainly applies deformable self-attention mechanism, and only performs self-attention operation on key points, which greatly reduces the computational overhead and space complexity.

\textbf{UNETR}.
UNEt TRansformers (UNETR) \cite{hatamizadeh2021unetr} is proposed for medical image segmentation, which is inspired by the Transformer in remote sequence learning. Pure Transformer is adopted as an encoder to learn the input's representation and obtain the global multi-scale information.
The encoder is directly connected to the decoder through jump connections of different resolutions to calculate the final semantic segmentation output.
Experiments have been evaluated on medical segmentation decathlon (MSD) data set (volumetric brain tumor and spleen segmentation tasks on  MR and CT modes). UNETR shows good performance on the data set above.

\textbf{Swin-unet}.
%
Swin-unet \cite{cao2021swin} is a UNet-like medical image segmentation network based on pure Transformer. It uses Swin Transformer to build encoder, bottleneck and decoder. The tokenized image blocks are fed into the U-shaped en-decoder architecture based on Transformer through jump connection, which can learn local and global semantic features. Hierarchical Swin Transformer with shifted window is used as encoder to extract context features. A decoder with a patch extension layer based on the symmetric Swin Transformer is designed, and the spatial resolution of the feature map is restored through up-sampling. Experiments have been evaluated on multi-organ and heart segmentation tasks. It shows that Swin-unet performs better than full convolution networks, convolution networks and Transformer-based methods (i,e. TransUnet \cite{chen2021transunet}).

\textbf{TransBTS}.
TransBTS \cite{wang2021transbts} is a method that combines Transformer and 3D CNN for 3D MRI brain tumor segmentation. Different from TransUNet \cite{chen2021transunet}, it can process image slices at one time. The encoder uses 3D CNN to extract volume space feature map, which can effectively capture 3D context information. Meanwhile, it carefully modifies the mapped tokens feature map, and then sends tokens to Transformer for global feature modeling. The decoder utilizes the Transformer embedded function and performs progressive up sampling to predict the detailed segmentation graph. The experimental results show that TransBTS performed well on the BRATS 2019 data set (brain tumor segmentation) than other methods.

\textbf{TransUNet}.
For modeling remote dependencies, TransUNet \cite{chen2021transunet}
combines the Transformer and UNet's advantages for medical image segmentation tasks. The Transformer encodes the labeled image blocks in the convolutional neural network (CNN) feature map to extract the global context. The decoder up-samples the encoded features and combines them with high-resolution CNN feature maps to achieve precise positioning. TransUNet has achieved better performance on multi-organ and heart segmentation than various competitive methods.

\textbf{TransClaw U-Net}.
TransClaw U-Net \cite{chang2021transclaw} combines convolution and Transformer operations in the coding part. Convolution is used to extract the shallow spatial features of the image, and the image resolution can be restored after up sampling. The Transformer operation can obtain global information between batches with different codes. The bottom-up sampling structure of the decoder can preserve the detail information well, so it can segment the detail information better. In contrast, the performance of TransClaw U-Net is better than other network structures in synapse multi organ segmentation data set.

\textbf{ViTBIS}.
Vision Transformer for biomedical image segmentation (ViTBIS) is proposed in \cite{cao2021swin}. It mainly performs $1\times 1$, $3\times 3$, and $5\times 5$ convolution on the input feature map to obtain multi-scale features before inputting into the encoder and decoder. Subsequently, the multi-scale feature maps are cascaded through the \textbf{Concat} operator and sent to three consecutive Transformer blocks. The Transformer's encoder-decoder blocks are connected by skip connections. Similarly, the Transformer block and multi-scale architecture are adopted in the decoder before the linear mapping generates the output segmentation map. ViTBIS has been tested on the Synapse multi-organ segmentation data set, automatic heart diagnosis challenge data set, brain tumor MRI segmentation data set, and spleen CT segmentation data set. It is superior to most CNN-based Transformer models.

\textbf{LiteTrans}.
To relieve the problem of low computational complexity, Transformer cannot perform more accurate segmentation of complex and low-contrast anatomical structures. LiteTrans \cite{xu2021litetrans} proposes that Transformer and CNN are deeply integrated in an Encoder-Decoder with skip-connection U-shaped architecture. It merges a new multi-branch module with convolution operation and local global self-attention (LGSA) into LiteTrans, bringing local and non-local feature interaction. It is a global self-attention approximation scheme with low computational complexity. Experiments show that it owns relatively few parameters and low calculations while ensuring accurate performance.

\textbf{nnFormer}.
Unlike previous frameworks that combine self-attention and convolution, nnFormer \cite{zhou2021nnformer} focuses on studying how to best combine self-attention and convolution. It learns volume representation from 3D partial volumes. Compared with simple voxel-level self-attention implementations, this volume-based operation helps reduce the computational complexity on Synapse and ACDC datasets by approximately 98\% and 99.5\%, respectively.

\textbf{AFTer-Unet}.
The existing 2D method models mainly use pure-Transformer to directly replace the convolutional layer or use Transformer as an additional encoder between the encoder and decoder of U-Net. However, these methods only consider single-slice attention coding, and do not use the axis-axis information provided by 3D voxels. In 3D settings, volume data and Transformer GPU memory consumption is huge, and down-sampling or post-cropping processing are of limitations. To solve the limitations above, AFTer-Unet \cite{yan2021after} is proposed. It takes full advantage of the ability of the convolutional layer to extract detailed features and Transformer strength for long sequence modeling. It also considers remote cues within and between slices to guide segmentation. At the same time, it has fewer parameters and less GPU memory to train than previous Transformer-based models.

\textbf{MedT}.
The local-global training strategy (LoGo) is proposed in \cite{valanarasu2021medical}, which can effectively train the model and further improve the performance on medical images segmentation. The model MedT operates on the entire image and patch to learn local and global features respectively. Besides, the gated Axial-Attention model is proposed to extend the existing architecture by introducing additional control mechanisms in the self-attention module.

\subsubsection{Remarks}
Transformer has also made great progress in image segmentation. Whether it is semantic segmentation or instance segmentation in natural images, the image segmentation model based on Transformer can show superior performance. This section focuses on investigating the application of Transformer in medical image segmentation. U-Net \cite{DBLP:journals/corr/RonnebergerFB15} is a common model with superior performance in image segmentation. The medical image segmentation models based on Transformer are mostly combined with U-Net model, such as UNETR \cite{hatamizadeh2021unetr}, SwinUnet \cite{cao2021swin}, TransUNet \cite{chen2021transunet}, TransClaw UNet \cite{chang2021transclaw}, and AFTer-Unet \cite{yan2021after}.
We counted the Transformer-based medical image segmentation model, and mainly compared its performance on the Synapse multi-organ CT data set, as shown in TABLE. \ref{tab:ct}. nnFormer achieved 87.4\% DSC, achieving a sate-of-art performance. There are many Transformer-based researches for image segmentation, especially for medical image. How to combine the natural image segmentation model with superior performance with the Transformer model may become a research direction.
\section{Transformer for video tasks}\label{sec:5}
Transformer has been developed in video learning and understanding, including object tracking, video classification, and video segmentation. In this section, Transformer-based methods for object tracking and video classification are mainly investigated.

\subsection{Object Tracking}
\begin{figure}[htp]
\setlength{\belowcaptionskip}{-0.1cm}
   \includegraphics[width=1.0\linewidth]{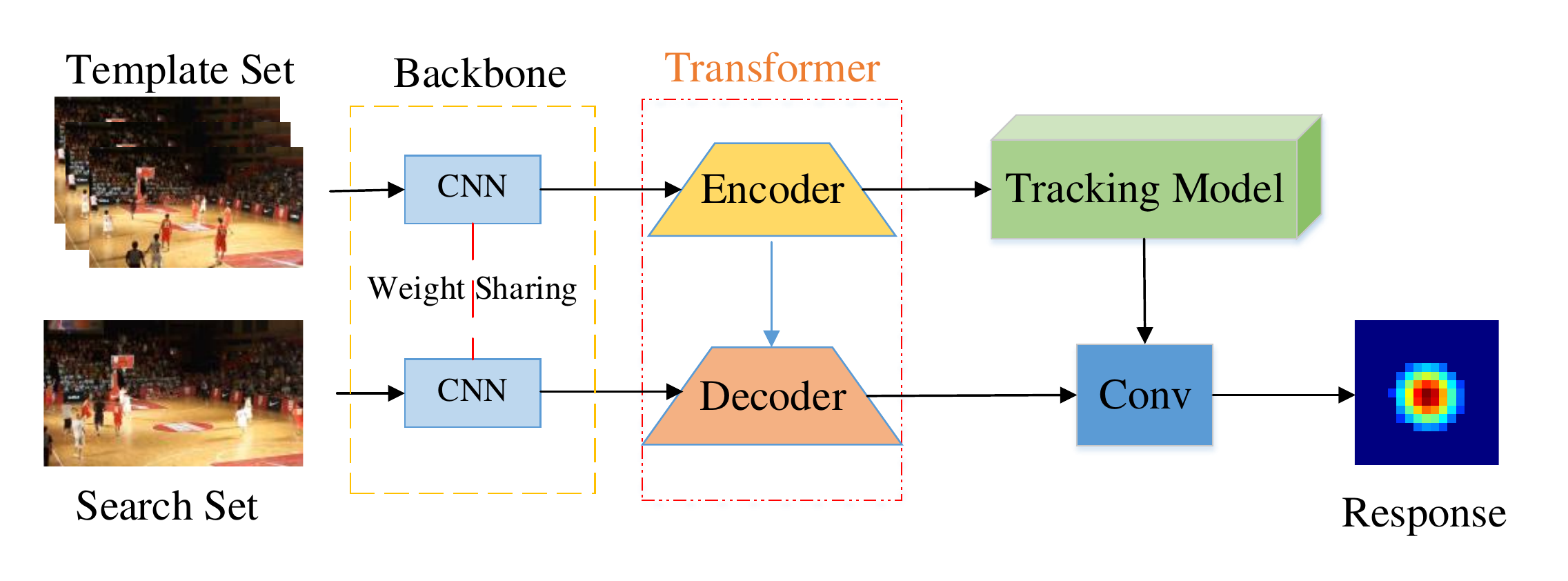}
   \caption{The basic framework of Transformer-assisted methods for object tracking.}
\label{fig:tracker}
\end{figure}

Transformer-based target tracking methods have gradually emerged in recent years. The basic framework of Transformer-assisted methods for object tracking is shown as Fig. \ref{fig:tracker}. The template and search are both sent to CNN backbone for feature extraction. After the features are extracted, the two-way features are sent to the two parallel branches of the quasi-Siamese network composed of the Transformer encoder and the decoder. The tracking model and the convolution convolve the encoded features of the template frame and the decoded features of the search frame respectively to obtain the response position of the object in the search frame.
Here, some typical single-object and multi-object tracking methods based on Transformer are introduced.
\begin{table*}[]
\centering
\caption{Different Transformer-based methods for multi-object tracking on MOT data sets, including MOT15, MOT17, MOT19, and MOT20.}
\begin{tabular}{c|c|c|c|c|c|c|c|c}
\toprule
\multicolumn{1}{l|}{\textbf{Methods}}                      & \textbf{Dataset}                & \textbf{MOTA$\uparrow$} & \textbf{IDF1$\uparrow$} & \textbf{MT$\uparrow$}  & \textbf{ML$\downarrow$} & \textbf{FP$\downarrow$}   & \textbf{FN$\downarrow$}   & \textbf{ID Sw$\downarrow$} \\ \toprule
\multicolumn{1}{l|}{TrackFormer(Private)\cite{meinhardt2021trackformer}} & \multirow{2}{*}{MOT17} & 65    & 63.9  & 1074 & 324  & 70443 & 123552 & 3528   \\ \cline{1-1} \cline{3-9}
\multicolumn{1}{l|}{TrackFormer(Pubulic)\cite{meinhardt2021trackformer}} &                        & 62.5  & 60.7  & 702  & 632  & 32828 & 174921 & 3917   \\ \hline
\multirow{2}{*}{MOTR\cite{chang2021transclaw}}                      & MOT16                  & 66.8  & 67    & 34.1 & 25.7 & 10364 & 49582  & 586    \\ \cline{2-9}
                                                  & MOT17                  & 67.4  & 67    & 34.6 & 24.5 & 32355 & 149400 & 1992   \\ \hline
TransTrack \cite{sun2020transtrack}                            & MOT17                  & 74.5  & 63.9  & 46.8 & 11.3 & 28323 & 112137 & 3663   \\ \hline
\multirow{4}{*}{TransMOT \cite{chu2021transmot}}                  & MOT15                  & 57    & 66    & 64.5 & 17.8 & 12454 & 13725  & 244    \\ \cline{2-9}
                                                  & MOT16                  & 76.7  & 76.8  & 56.5 & 19.2 & 14999 & 26967  & 517    \\ \cline{2-9}
                                                  & MOT17                  & 68.7  & 72.2  & 33.5 & 31   & 8070  & 167602 & 1014   \\ \cline{2-9}
                                                  & MOT20                  & 73.1  & 74.3  & 54.3 & 14.6 & 12366 & 125665 & 1042   \\ \hline
ViTT \cite{zhu2021vitt}                                             & MOT16                  & 65.7  & 66.5  & 39.5 & 20.6 & -  & -   & -   \\ \hline
\multirow{2}{*}{TrackCenter \cite{xu2021transcenter}}                      & MOT17                  & 68.8  & 61.4  & 36.8 & 23.9 & 22.86 & 149188 & 4102   \\ \cline{2-9}

\label{tab:4}                                                 & MOT20                  & 61    & 49.8  & 48.4 & 15.5 & 49189 & 147890 & 4493   \\ \hline
\end{tabular}
\end{table*}

\subsubsection{Single object tracking}
The Transformer-based methods for single-object tracking mainly includes TransT \cite{chen2021Transformer}, STARK \cite{yan2021learning}, and SwinTrack \cite{lin2021swintrack}.

\textbf{TransT}.
TransT \cite{chen2021Transformer} is a single-object tracking method based on Transformer, which is mainly composed of the backbone, feature fusion network and the prediction head. The algorithm pipeline is as follows. The backbone network extracts the features of template and search area respectively. And then, the extracted features are sent to the feature fusion network after reshape to obtain the feature vector. Finally, the feature head performs regression prediction on the input feature vector to acquire the tracking result. Based on Transformer, TransT combines ego context augmentation (ECA) module to better integrate the correlation between template and candidate features. Therefore, TransT can obtain better tracking results and achieve real-time tracking speed.

\textbf{STARK}.
STARK \cite{yan2021learning} was proposed in CVPR in 2021. It is a visual tracker based on the Transformer's encoder-decoder structure. It is inspired that Transformer is globally dependent and can mine the time and space in target tracking. Encoder, decoder, and prediction header are three essential parts of STARK.
%
For the STARK, the backbone extracts the features of the input initial target object and a dynamically updated template object.
The extracted features are flattened and jointly input into the Transformer's encoder. Transformer can learn robust spatiotemporal joint representation in the whole process, and its encoder can capture the spatial and temporal information of the target simultaneously. The encoded features and label query are sent to the decoder for decoding at the same time, and then the tracking target position is obtained through the bounding box prediction header. STARK transforms target tracking into a direct bounding box prediction problem without using any proposal or predefined anchor. Meanwhile, the end-to-end stark simplifies the tracking process. It has achieved excellent results in long-term and short-term tracking benchmarks, and can obtain real-time tracking speed.

\textbf{SwinTrack}.
Swin-Transformer Tracker, namely SwinTrack, is proposed in \cite{lin2021swintrack} based entirely on attention. It also allows full interaction between the target object and the search area for tracking. It mainly includes Swin Transformer backbone, serial-based fusion encoder, and general position-coding solution. Strategies such as feature fusion, location coding and training loss can effectively improve the performance of SwinTrack. Experiments have been evaluated on LaSOT, TrackingNet, and GOT-10k data sets. It is surpassing STARK by 3.1\% at 45 FPS on the LaSOT benchmark.

\subsubsection{Multi-object tracking}
Transformer-based tracking methods, including TransTrack \cite{sun2020transtrack}, TrackFormer \cite{meinhardt2021trackformer}, TrSiam and TrDiMP \cite{wang2021Transformer}, ViTT \cite{zhu2021vitt},
TransCenter \cite{xu2021transcenter} are introduced here.

\textbf{TransTrack}.
TransTrack \cite{sun2020transtrack} is constructed based on the query-key architecture. In the initial frame, the key is obtained after the backbone network extracts the features. The query comes from the target feature query set of the previous frame and a learnable target query set. The idea of detecting branch learned object query comes from DETR, a learnable representation that can learn to query the target's location from the key to complete the detection. The detection branch completes all targets on the current frame to obtain detection boxes. In the tracking frame branch, the object feature query is the target's feature vector of the previous frame generated by the detection branch. The object feature query queries the target's position in the current structure from the key, performs displacement prediction, and finally obtains the tracking boxes. Finally,  the tracking can be completed by simple IOU matching after the current frame's tracking frame and detection frame are accepted.

\textbf{TrackFormer}.
TrackFormer \cite{meinhardt2021trackformer} is a tracker that applies Transformer to multi-object tracking (MOT). It is a novel MOT framework, which introduces Transformer into MOT and completes the migration from detection to tracking with DETR through the design of track query. It treats MOT as a set of prediction problems and conducts joint detection and tracking through attention. First, the CNN backbone still extracts the features of the initial frame. And then, the extracted features are forwarded to the Transformer encoder. The learning target query is under the encoder, which can query the corresponding number of output embeddings, generating bounding boxes and output embedding of category information. The successfully predicted target is sent to the next frame as a tracking query. It is sufficient to perform DETR detection on the initial and subsequent frames. The decoder detects the current frame according to the track and object queries. If the confidence of the frame obtained by the object query is higher than the threshold, it is considered that a new target is generated. If the confidence of the track query is less than the threshold, the trajectory is considered to be terminated.

\textbf{TrSiam \& TrDiMP}. TrSiam and TrDiMP models are proposed in \cite{wang2021Transformer}, which specifically combines Transformer with the latest discriminant tracking pipeline. It mainly proposes to utilize the Transformer to assist in tracking the frame of video frames for fixed object tracking, which can alleviate the problem of ignoring the context of existing video trackers. TrSiam model divides the encoder and decoder of the Transformer into two parallel branches and carefully designs them in a tracking pipeline similar to Siamese. The encoder promotes the target template through attention-based feature enhancement, conducive to the generation of high-quality tracking models. The decoder propagates the tracking hint from the previous template to the current frame, simplifying the object search process. The proposed models are verified on the public data set and they show relatively excellent performance.

\textbf{ViTT}.
Vision Transformer Tracker \cite{zhu2021vitt}, namely ViTT, is also a Transformer-based model for multi-target tracking.
Transformer encoders are adopted as a backbone network, which models the global context through each encoder to solve occlusion challenges and complex scenes. The model adopts multi-task learning to simultaneously output the object positions and their corresponding appearance embeddings in a shared network. Experiments show that it reaches 65.7\% on the MOT16 data set.

\textbf{TrackCenter}.
There is an incompatibility problem between the Transformer structure and the bounding box representation of the multi-target tracking task, which is not suitable for Transformer learning. To solve the situation above, TransCenter \cite{xu2021transcenter} is proposed as a Transformer-based MOT framework that performs multi-target tracking in the form of centers. Using a dense query in the dual-decoder network can reliably infer the target center heatmap and correlate it over time. The experiment verified the proposed model's performance. And it achieves 68.8\% and 61\% accuracy on MOT17 and MOT20 the respectively.

\begin{table*}[]
\centering
\caption{Comparison of Transformer-based Methods on Kinetics 400 for video classification. Top-1.acc (\%) and top-5.acc (\%) accuracy on the testing data. For ``views", $x\times y$ means $x$ temporal crops and $y$ spatial crops. ``FE" represents the Factorised Encoder model.}
\label{tab:400}
\begin{tabular}{c|c|c|c|c|c|c}
\toprule
\textbf{Methods}        & \textbf{Pre-training data} & \textbf{Top-1.acc} & \textbf{Top-5.acc} & \textbf{Views} & \textbf{\#TFLOPs}    & \textbf{Para(M)} \\ \toprule
TimeSformer \cite{bertasius2021space}                           & ImageNet-21K     & 78    & 93.7  & -  & 0.59        & 121.4   \\ \hline
TimeSformer \cite{bertasius2021space}                          & -       & 80.7  & 94.7  & $1\times 3$ & 2.38        & 121     \\ \hline
TimeSformer-HR \cite{bertasius2021space}                       & ImageNet-21K     & 79.7  & 94.4  & -  & 5.11        & -    \\ \hline
TimeSformer-L \cite{bertasius2021space}                        & ImageNet-21K     & 80.7  & 94.7  & $1\times 3$ & 7.14        & -    \\ \hline
ViViT-L/16x2 FE \cite{arnab2021vivit}                      & -       & 80.6  & 92.7  & $1\times 3$ & 3.98        & -    \\ \hline
ViViT-L/16x2 FE \cite{arnab2021vivit}                      & -       & 81.7  & 93.8  & $1\times 3$ & 11.94       & -    \\ \hline
ViViT-L/16x2 FE \cite{arnab2021vivit}                      & JFT        & 83.5  & 94.3  & $1\times 3$ & 11.94       & -    \\ \hline
ViViT-H/14x2 \cite{arnab2021vivit}                         & JFT        & 84.9  & 95.8  & $4\times 3$ & 47.77       & -    \\ \hline
R50-VTN \cite{neimark2021video}                              & ImageNet-1K         & 71.2  & 90    & -  & -        & -    \\ \hline
R101-VTN \cite{neimark2021video}                             & ImageNet-1K         & 72.1  & 90.3  & -  & -        & -    \\ \hline
DeiT-B-VTN \cite{neimark2021video}                           & ImageNet-1K         & 75.5  & 92.2  & -  & -        & -    \\ \hline
DeiT-BD-VTN \cite{neimark2021video}                          & ImageNet-21K     & 75.6  & 92.4  & -  & -        & -    \\ \hline
ViT-B-VTN \cite{neimark2021video}                            & ImageNet-21K     & 78.6  & 93.7  & -  & -        & -    \\ \hline
ViT-B-VTN+ \cite{neimark2021video}                           & ImageNet-21K     & 79.8  & 94.2  & -  & -        & -    \\ \hline
SCT-S \cite{zha2021shifted}                                & -       & 78.4  & 93.8  & $4\times 3$ & 0.088       & 19      \\ \hline
SCT-M \cite{zha2021shifted}                                & -       & 81.3  & 94.5  & $4\times 3$ & 0.163       & 33      \\ \hline
SCT-L \cite{zha2021shifted}                                & -       & 83    & 95.4  & $4\times 3$ & 0.343       & 60      \\ \hline
MViT-B 64x3 \cite{li2021improved}                          & -       & 81.2  & 95.1  & $3\times 3$ &0.455       & 37      \\ \hline
ViViT-L \cite{arnab2021vivit}                              & -       & 81.3  & 94.7  & $4\times 3$ & 0.3992       & 89      \\ \hline
MViT-B 64x3 \cite{li2021improved}                          & -       & 82.9  & 95.7  & $1\times 5$ & 0.225  & 51.2    \\ \hline
MViT-L↑ 312 \textasciicircum{}2 , 40×3 \cite{li2021improved} & ImageNet-21K     &\textbf{86.1}  & \textbf{97.0}  & $3\times 5$   & 2.828 & 217.6   \\ \hline
\end{tabular}
\end{table*}
\subsubsection{Remarks}
Vision Transformer has also been developed in the field of target tracking. Occlusion is still a challenge for target tracking of deep learning. It has been proven effective in relieving the occlusion problem and owns stronger robustness. For single-target and multi-target tracking, different Transformer-based models are proposed. The performance of the Transformer-based models for single object tracking has been evaluated on the public tracking benchmark, including VOT2018, VOT2019, OTB-100, UAV, NFS, TrackingNet, and LaSOT. Here, the
Transformer-based models for multi-object tracking and their performance on MOT series data sets are counted as TABLE \ref{tab:4}.
By contrast, there are still few related models. Thus, developing related Transformer models will become a research hotspot.
%
\subsection{Video classification}
\begin{table*}[htp]
\centering
\caption{Comparison of Transformer-based Methods on Kinetics 600 for video classification. Top-1.acc (\%) and top-5.acc (\%) accuracy on the testing data. For ``views", $x\times y$ means $x$ temporal crops and $y$ spatial crops. ``FE" represents the Factorised Encoder model.}
\label{tab:600}
\begin{tabular}{c|c|c|c|c|c|c}
\toprule
\textbf{Methods}                    & \textbf{Pre-training data} & \textbf{Top-1.acc(\%)} & \textbf{Top-5.acc(\%)} & \textbf{Views} & \textbf{\#TFLOPs }   & \textbf{Para(M)} \\ \toprule
TimeSformer \cite{bertasius2021space}                           & ImageNet-21K     & 79.1  & 94.4  & -  & -        & -    \\ \hline
TimeSformer \cite{bertasius2021space}                          & -       & 82.4  & 96    & $1\times 3$   & 121         & -    \\ \hline
TimeSformer-HR \cite{bertasius2021space}                       & ImageNet-21K     & 81.8  & 95.8  & -  & -        & -    \\ \hline
TimeSformer-L  \cite{bertasius2021space}                       & ImageNet-21K     & 82.2  & 95.6  & -  & -        & -    \\ \hline
ViViT-L/16x2 FE \cite{arnab2021vivit}                       & -       & 82.9  & 94.6  & -  & -        & -    \\ \hline
ViViT-L/16x2 FE \cite{arnab2021vivit}                       & JFT        & 84.3  & 94.9  & -  & -        & -    \\ \hline
ViViT-H/14x2 \cite{arnab2021vivit}                          & JFT        & 85.8  & 96.5  & -  & -        & -    \\ \hline
SCT-S \cite{zha2021shifted}                                & -       & 77.5  & 93.1  & -  & -        & -    \\ \hline
SCT-M \cite{zha2021shifted}                                & -       & 81.7  & 95.5  & $4\times 3$ & 33          & -    \\ \hline
SCT-L \cite{zha2021shifted}                                & -       & 84.3  & 96.3  & $4\times 3$ & 60          & -    \\ \hline
ViViT-L \cite{arnab2021vivit}                              & -       & 83    & 95.7  & $4\times 3$   & 89          & -    \\ \hline
MViT-B 64x3 \cite{li2021improved}                           & -       & 85.5  & 97.2  & $1\times 5$ & 0.206  & 51.4    \\ \hline
MViT-L↑ 312\textasciicircum{}2 , 40×3 \cite{li2021improved} & ImageNet-21K     & \textbf{87.9}  & \textbf{97.9}  & $3\times 4$   & 3.790 & 217.6   \\ \hline
\end{tabular}
\end{table*}

\begin{figure}[htp]
\setlength{\belowcaptionskip}{-0.1cm}
   \includegraphics[width=1.0\linewidth]{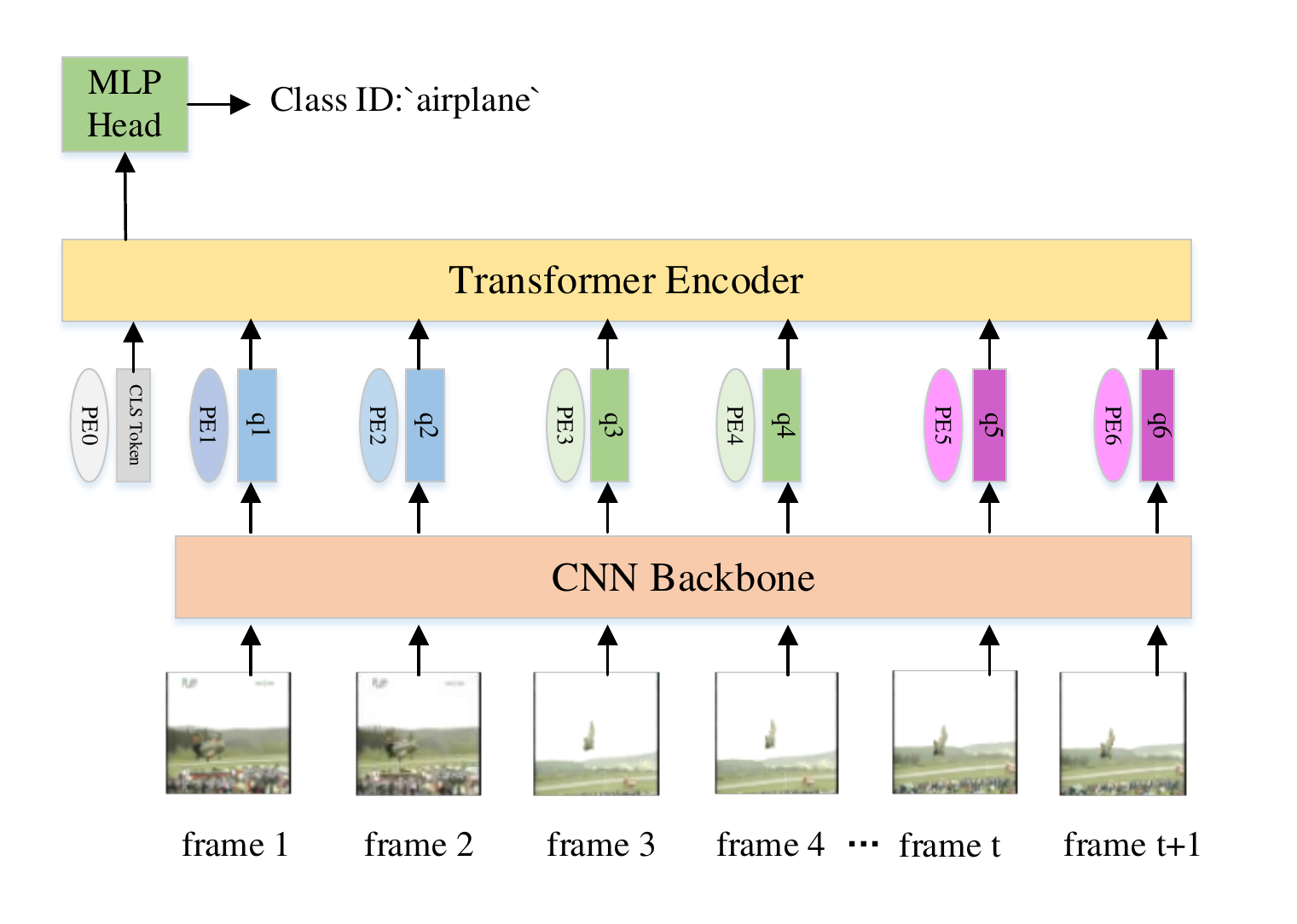}
   \caption{The basic framework of Transformer-assisted methods for video classification. `$q_{i}$' means the feature extracted by CNN of the $frame_{i}$. $PE_{i}$ represents the position encoding of the $frame_{i}$.}
\label{fig:vc}
\end{figure}

Video learning and understanding play essential roles in today's multimedia applications. With the emergence of video information, video classification is also a research hotspot with excellent research and commercial value. The basic framework of Transformer-assisted methods for video classification is shown as Fig. \ref{fig:vc}. Different frames of a video are sent to CNN to extract their features. And then, the extracted features are sent to the Transformer encoder. Finally, the encoded features are forwarded to the MLP head to predict the label of the input video. Here, Transformer-based methods for video classification are investigated.


\textbf{VTN}.
Unlike traditional 3D ConvNets, VTN \cite{neimark2021video} is proposed to classify actions by participating in the entire video sequence information. Its structure comprises three parts,  including a 2D spatial backbone network for feature extraction. Longformer is followed by an encoder of temporal attention, which uses feature vectors combined with positional encoding. Subsequently, the [CLS] token is processed by the classification MLP header to obtain the final class prediction. Built on any given 2D space, VTN is a general-purpose network with superior performance and competitive accuracy. The results of its competition in Kinetics-400 and Moments are on the time base and current ablation studies.

\textbf{TimeSformer}.
TimeSformer \cite{bertasius2021space} is a non-convolutional video classification method based on self-attention. It adapts the standard Transformer architecture to video by directly learning the spatio-temporal features from the frame-level patch sequence. The experiment applies temporal and spatial attention in each block to obtain the best video classification accuracy in the considered design choices. It achieved 82.2\% and 80.7\% accuracy in Kinetics-400 and Kinetics-600, respectively.

\textbf{MViT-L}.
MViT \cite{li2021improved} is a multi-scale visual Transformer, a unified architecture for image and video classification and object detection. It proposes an improved version of MViT that includes decomposed relative position embedding and residual pool connection. MViT has the most advanced performance in 3 areas: ImageNet classification accuracy rate is 88.8\%, COCO target detection accuracy rate is 56.1 APbox, and Kinetics-400 video classification accuracy is 86.1\%.

\textbf{ViViT}. ViViT \cite{arnab2021vivit} is a video vision Transformer, which performs well in video classification. In ViViT, it extracts the spatiotemporal tokens from the input video. And then, the extracted tokens are encoded by the Transformer layers. Meanwhile, several variants are proposed to handle the long sequences of tokens encountered in the video. ViViT shows state-of-the-art results on several video classification benchmarks, which is prior to deep 3D convolutional networks.


\textbf{VidTr}.
Video Transformer (VidTr) \cite{zhang2021vidtr} is a Transformer without convolutions for video classification. It can aggregate spatio--temporal information via stacked attention. Compared with 3D CNNs, it performed better with higher efficiency. VidTr can reduce the memory cost of the vanilla video Transformer. The standard deviation-based top-K pooling for attention is proposed for optimizing the model and reducing the computation. VidTr has been evaluated on five commonly used data sets. It shows a good balance efficiency and effectiveness. The visualization shows that VidTr performs well on predicting actions that require long-term temporal reasoning.

\textbf{TokShift Transformer}.
TokShift Transformer \cite{zhang2021token} is a pure convolutional-free video Transformer pilot with computational efficiency for video understanding. The proposed Token Shift Module uses zero-parameter and zero-FLOPs, which models temporal relations within each Transformer encoder. It barely temporally shifts partially [Class] token features back-and-forth across adjacent frames. Besides, it is plugged into each encoder of a plain 2D vision Transformer. Experiments show its robustness, effectiveness, and efficiency on several benchmark data set for video understanding.

\textbf{Video Swin Transformer}.
The video swin Transformer is proposed in \cite{liu2021video} based on the Swin Transformer. The Transformer layers can globally connect patches across the spatial and temporal dimensions. The inductive bias of locality is introduced into the video Transformer, which obtains a trade-off of speed and accuracy. The locality of the Video Swin Transformer is realized by adapting the Swin Transformer designed for the image domain. Experiments show 84.9\%, 86.1\%, and 69.6\% top-1 accuracy on Kinetics-400, Kinetics-600, and Something-Something v2.

\subsection{Remarks}
Transformer has relatively many applications in video classification.
This paper counts the performance of related models on the video classification data sets KINETICS 400 \cite{kay2017kinetics} and KINETICS 600 \cite{carreira2018short}, as shown in TABLE \ref{tab:400} and TABLE \ref{tab:600}. MViT-L achieved 86.1\% top-1 accuracy on the data set KINETICS 400, and 87.9\% top-1 accuracy on the data set KINETICS 600, which is the model with the most superior performance among the existing models. Traditional video classification models use 3D convolution filters. However, such filters effectively capture short-range modes in local temporal and spatial regions, but they cannot model the temporal and spatial dependencies beyond their acceptance domain. The Transformer's self-attention mechanism can capture the temporal and spatial dependencies of the entire video. Therefore, Transformer will continue to develop in the task of video classification. Besides,
the Action Transformer model \cite{girdhar2019video} is proposed for recognizing and localizing human actions in video clips. It raises our attention to other more video understanding application tasks.

\section{Ten public issues and conclusion}\label{sec:6}
In this section, the ten public issues of Transformer-based methods are listed, and the conclusion of this survey is given.
\subsection{Ten Public Issues}
Transformer can directly calculate the correlation between each word without passing through a hidden layer, and it can also perform parallel calculations to make full use of GPU resources. Meanwhile, Transformer has made specific progress in existing deep learning applications, and its applications are gradually expanding to visual learning understanding. Of course, there are still some limitations in its application development. Ten public issues related to Transformer-based research are summarized as follows.

1) The ability to acquire partial information is weak. Compared with CNN and RNN, the ability of the Transformer to obtain the local feature is relatively weak. Improving or enhancing the Transformer's local expression ability may become a significant breakthrough.
Besides, there are many advantages of CNNs. Combining the CNNs thought with Transformer can improve the Transformer's local expression ability. For example, HRViT \cite{gu2021hrvit} is proposed by combining HRNet \cite{wang2020deep} with ViT. It shows that combining multi-scale CNN and Transformer has certain reference significance for our subsequent improvement of Transformer.

2) The location information encoding problem. Using word vectors can preserve the semantics of the word vector by a linear transformation. However, position-coding does not own this kind of transformability in the semantic space, and it is equivalent to a type of artificially designed index. Then, it is unreasonable to add this position code to the word vector. Thus the position information cannot be well represented.

3) The top layer gradient disappears. The original Transformer model combines some residual modules and layer normalized LN modules. The residual error passes through the LN layer, resulting in a decrease in gradient. At the same time, there is no direct connection between the final output layer and the previous Transformer layer, and the layer normalization module will block the gradient flow. The gradient on the top layer will still disappear when multiplying multiple times.

4) The computing power requirements are high. Much Transformer-related research often requires many GPUs for experiments, which is unfavorable for most researchers. Reducing the number of Transformer calculations, simplifying its models, and speeding up its calculations may become a breakthrough in research, facilitating the further development of its theory.

5) The performance of Transformer-based methods still needs further improvement. CNN has been successful in many visual learning understanding tasks, but there are still some challenges. For example, the recognition rate cannot reach excellent face recognition accuracy and detection in the video tasks.

6) Adaption to computer vision tasks. The existing visual Transformer has done some preliminary explorations of adapting the Transformer structure in NLP to visual studies. In the future, designing Transformers that are more adapted to the characteristics of CV will bring better performance improvements.

7) The number of parameters and calculations is relatively large. Its space and time complexity are both $O(n^2)$ level, where $n$ is the sequence length. When $n$ is relatively large, the Transformer model's calculation is unbearable. The existing visual Transformer parameters and computations are extensive. For example, ViT requires 18B FLOPs to reach about 78\% top-1 accuracy on ImageNet, but CNN models such as GhostNet \cite{han2020ghostnet} only need 600M FLOPs to get more than 79\% top-1 accuracy. Therefore, an efficient Transformer for CV needs to be developed to be comparable to CNN.

8) Dependence on extensive pre-training data. The performance of many existing Transformers often requires specific pre-training strategies. For example, pre-training is often needed on ImageNet or even the undisclosed JF300M data set in classification tasks. However, an excellent pre-training model requires a massive cost of computing power, so how to train the Transformer model efficiently has also become a focus of attention. Meanwhile, designing transformer methods without pre-training is also a fundamental problem.

9) The application research of Transformer for image segmentation needs to be further expanded. Investigation and study show that the existing Transformer image segmentation is mainly carried out on medical image research, while the natural and remote sensing images are less researched. Therefore, we hope to see more universal Transformer segmentation models appear.

10) Transformer-based language model research can reference computer vision multi-task and multi-modal learning tasks.
Transformer originated from the field of NLP, and the focus on multi-tasking and multi-modal input methods of it in natural language learning will be a reference for more visual learning understanding researches. Therefore, effectively combining Transformer with multi-task and multi-modal tasks will become a significant research direction.
\subsection{Conclusion}
This survey comprehensively investigates the development of Transformer in visual learning understanding and gives some remarks.
Notably, some critical experimental performance statistics of Transformer-based methods are shown in several images and video tasks,
which provide researchers with the experiment performance reference. Meanwhile, ten open problems of Transformer-based models, such as complex calculation, weak local representation ability, and relying on a large number of pre-training data, are presented.
Of course, some developing directions are also given.
This survey aims to make researchers have a comprehensive understanding of Transformer-based researches, which has a significant meaning for promoting the development of Transformer.
\bibliographystyle{IEEEtran}
\bibliography{mybibtex}

\end{document}